\documentclass[10pt,twocolumn,letterpaper]{article}

\usepackage{iccv}
\usepackage{times}
\usepackage{epsfig}
\usepackage{graphicx}
\usepackage{amsmath}
\usepackage{amssymb}
\usepackage{textcomp}
\usepackage{multirow}
\usepackage[]{algorithm2e}
\usepackage{adjustbox}
\usepackage{array}
\usepackage{csquotes}
\usepackage{xpatch}
\usepackage{verbatim}
\usepackage{diagbox}
\usepackage{enumitem}

\DeclareMathOperator*{\argmin}{arg\,min}
\newcommand\np{n_\mathcal{P}}
\newcommand\p{\mathcal{P}}
\newcommand\pt{_{\mathsf{pt}}}
\newcommand\se{_{\mathsf{seg}}}
\newcommand\setwo{_{\mathsf{}2}}
\newcommand\co{_{\mathsf{co}}}
\newcommand\NFA{\mathit{NFA}}
\newcommand\la{\lambda}
\newcommand\hp{\hat{p}}
\newcommand\tp{\tilde{p}}
\newcommand\tP{\tilde{P}}

\newcommand\cl{\check{l}}
\newcommand\hl{\hat{l}}
\newcommand\tL{\tilde{L}}
\newcommand\tl{\tilde{l}}
\newcommand\trsp{\top}
\newcommand\tr{T}

\newcommand\td{t}

\newcommand{\smallsection}[1]{\smallskip\textbf{#1.}}

\usepackage[pagebackref=true,breaklinks=true,letterpaper=true,colorlinks,bookmarks=false]{hyperref}

\iccvfinalcopy 

\def\iccvPaperID{2988} 

\ificcvfinal\fi
\begin{document}


\title{Robust SfM with Little Image Overlap}


\author{Yohann Sala\"un$^{1,2}$, Renaud Marlet$^1$, and Pascal Monasse$^1$\\
$^1$LIGM, UMR 8049, \'Ecole des Ponts, UPE, Champs-sur-Marne, France\\
$^2$CentraleSup\'elec, Ch\^atenay-Malabry, France\\
{\tt\small \{yohann.salaun,renaud.marlet,pascal.monasse\}@imagine.enpc.fr}
}

\maketitle


\begin{abstract}
Usual Structure-from-Motion (SfM) techniques require at least trifocal overlaps to calibrate cameras and reconstruct a scene.  We consider here scenarios of reduced image sets with little overlap, possibly as low as two images at most seeing the same part of the scene.
We propose a new method, based on line coplanarity hypotheses, for estimating the relative scale of two independent bifocal calibrations sharing a camera, without the need of any trifocal information or Manhattan-world assumption.  We use it to compute SfM in a chain of up-to-scale relative motions.  For accuracy, we however also make use of trifocal information for line and/or point features, when present, relaxing usual trifocal constraints.  For robustness to wrong assumptions and mismatches, we embed all constraints in a parameterless RANSAC-like approach.  Experiments show that we can calibrate datasets that previously could not, and that this wider applicability does not come at the cost of inaccuracy.
\end{abstract}
\vspace{-4mm}

\section{Introduction}

Structure-from-Motion (SfM) has made spectacular improvements concerning scalability \cite{Frahm2010rome,Wu2013linsfm,Heinly2015cvpr}, accuracy \cite{moulon2013global,Cui2015bmvc,Arrigoni2015transnorm,Sweeney2015iccv} and robustness 
\cite{Wilson2014eccv,Ozyesil2015cvpr}.  However, most approaches assume a significant amount of overlap between images.
In this work, we consider the case where images have only little overlap, possibly as low as two images at most seeing the same part of the scene.

This situation commonly occurs when a scene is photographed by people with little or no knowledge in photogrammetry.  Even when informed, they can make occasional mistakes, widening too much a baseline, or shooting a low-quality image that cannot be exploited and has to be skipped.  Another example is when exploiting street views, e.g., taken from a vehicle, if the viewpoints are too distant one from another, or if the camera is too close to the facades because of narrow sidewalks.  In such cases, SfM methods may yield partial or fragmented calibrations.

It also applies to situations where only a small number of pictures can be shot, because of physical or time constraints. It may be the case when digitizing a building which is still in use, not to disturb occupants.  Reducing the number of images, to only a few per room for indoor scenes, is also a way to reduce the cost and time for acquiring and processing information, as long as a minimum level of accuracy can still be reached for the targeted application.

\begin{figure*}[!t]
\begin{center}
\begin{tabular}{ccc}
\includegraphics[width=0.65\columnwidth]{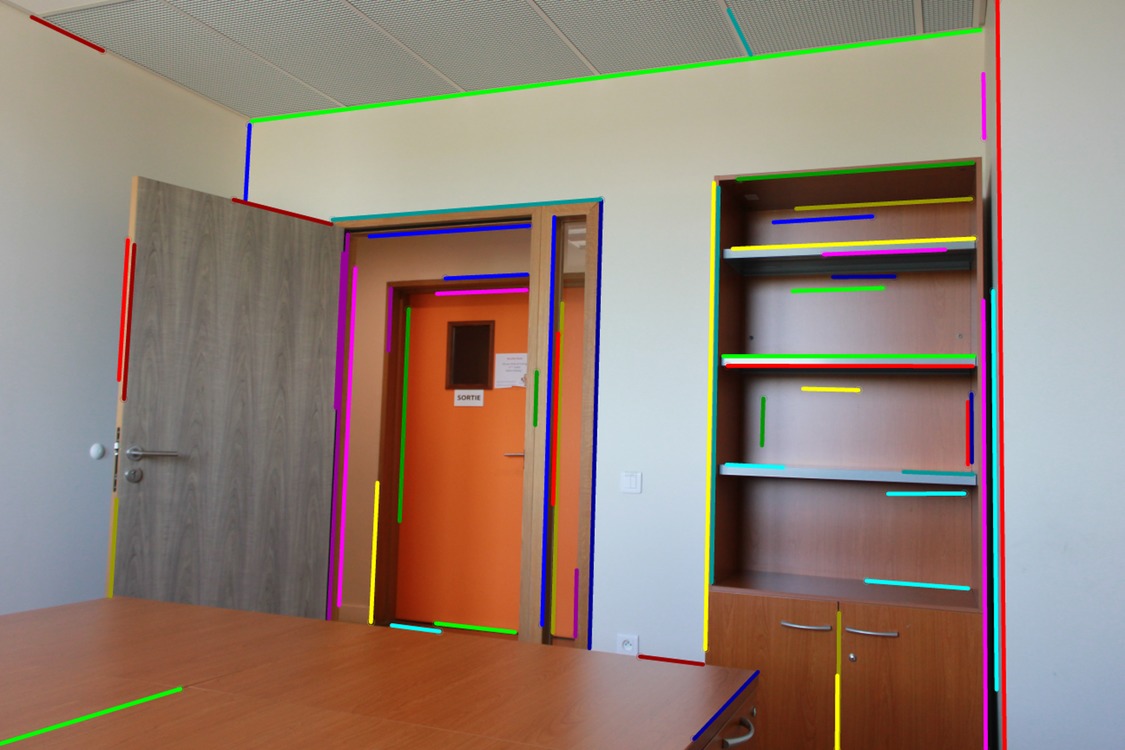} &
\includegraphics[width=0.65\columnwidth]{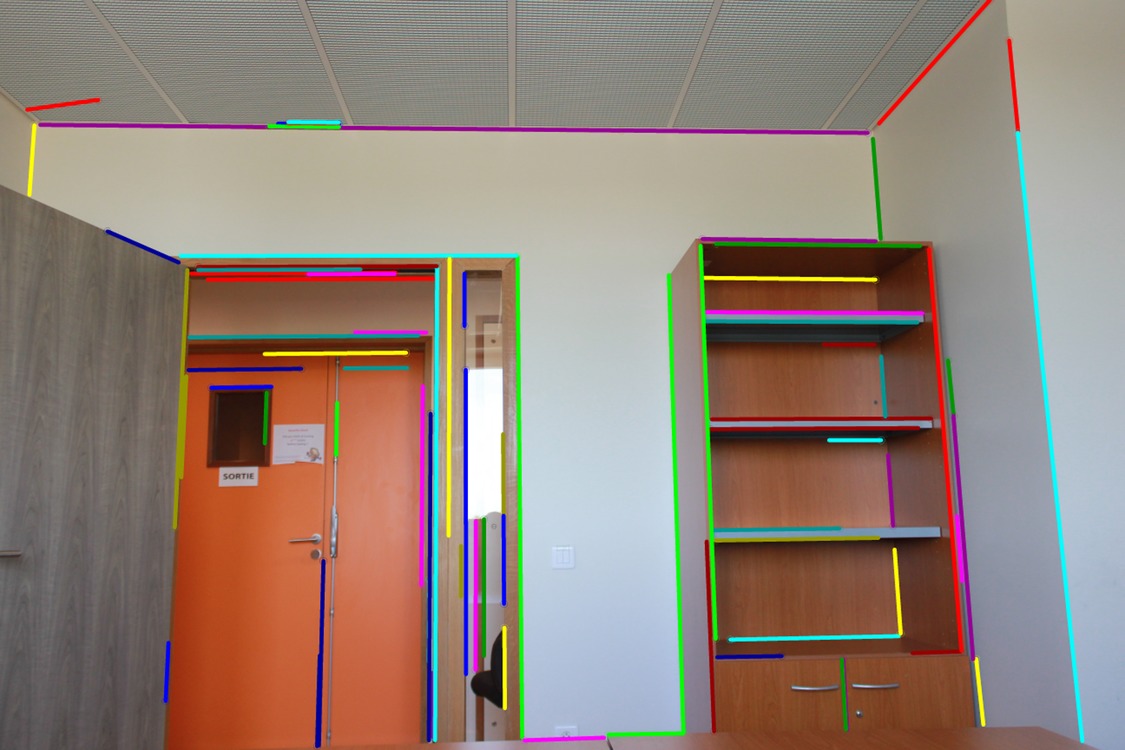} &
\includegraphics[width=0.65\columnwidth]{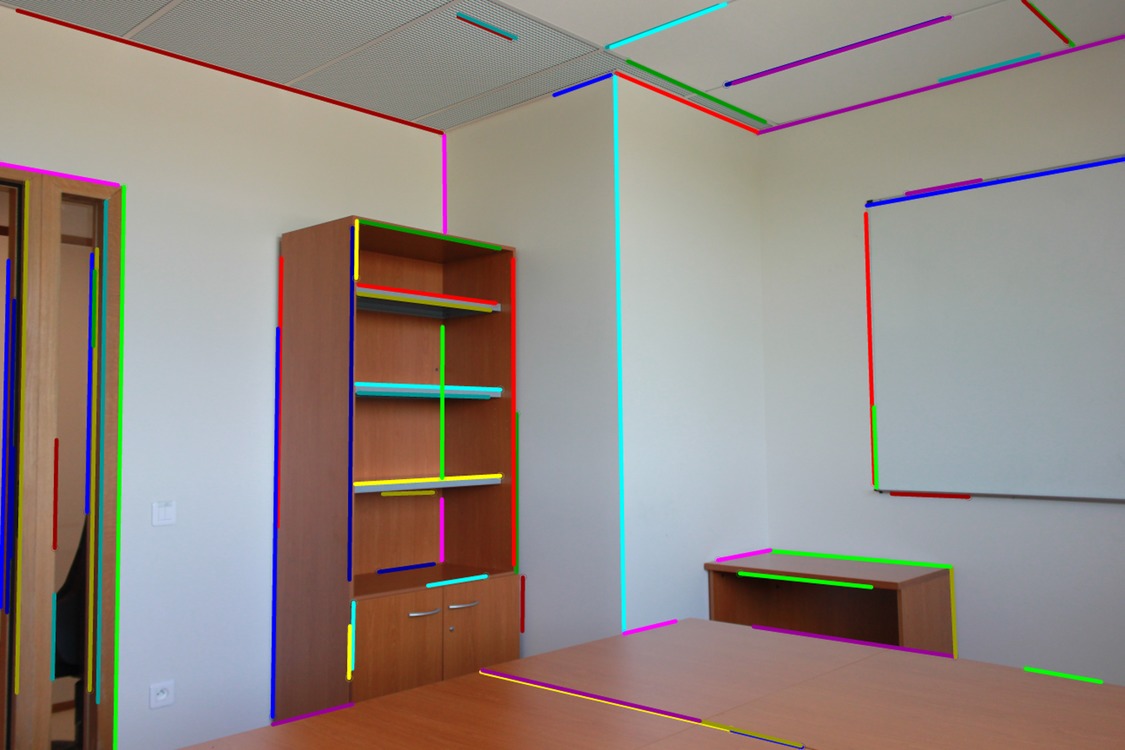}
\\
\includegraphics[width=0.65\columnwidth]{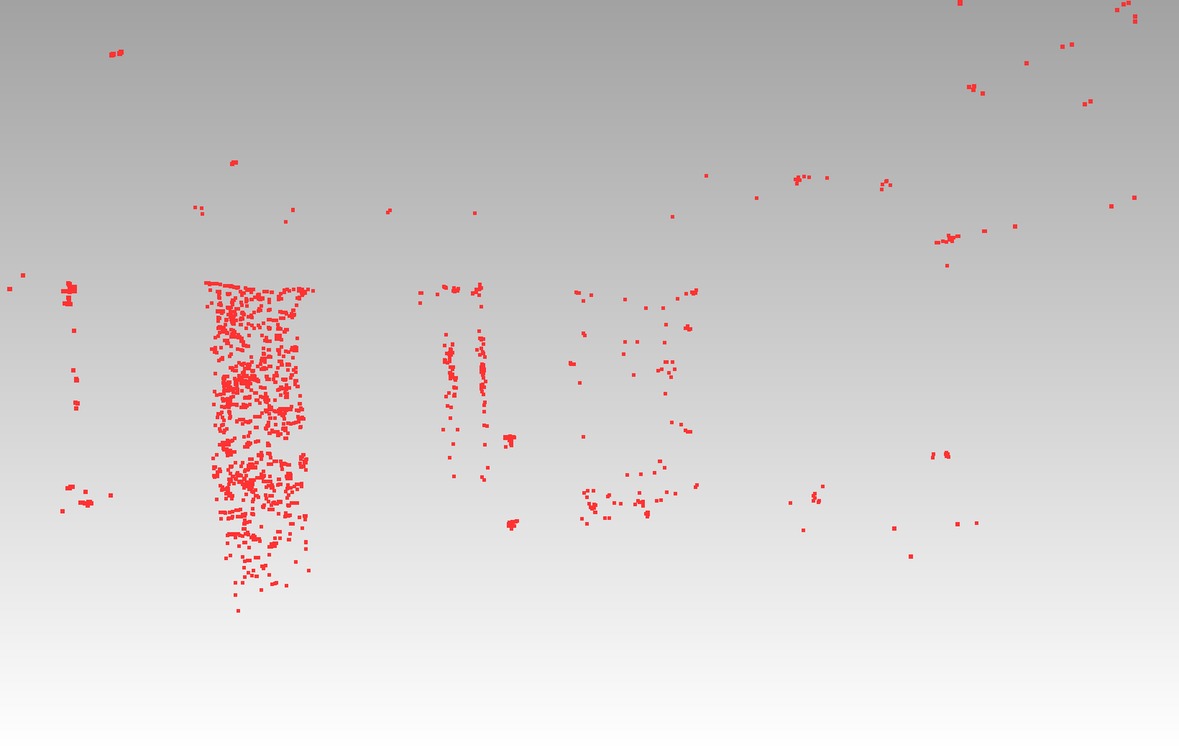} &
\includegraphics[width=0.65\columnwidth]{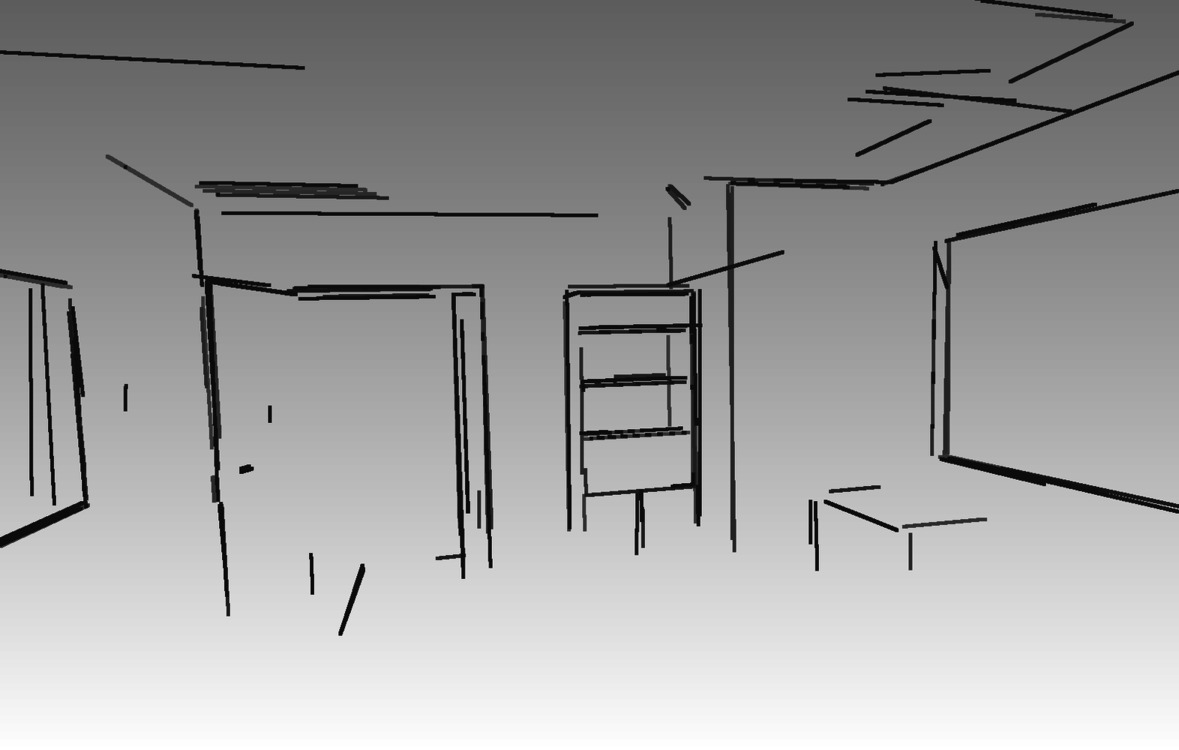} &
\includegraphics[width=0.65\columnwidth]{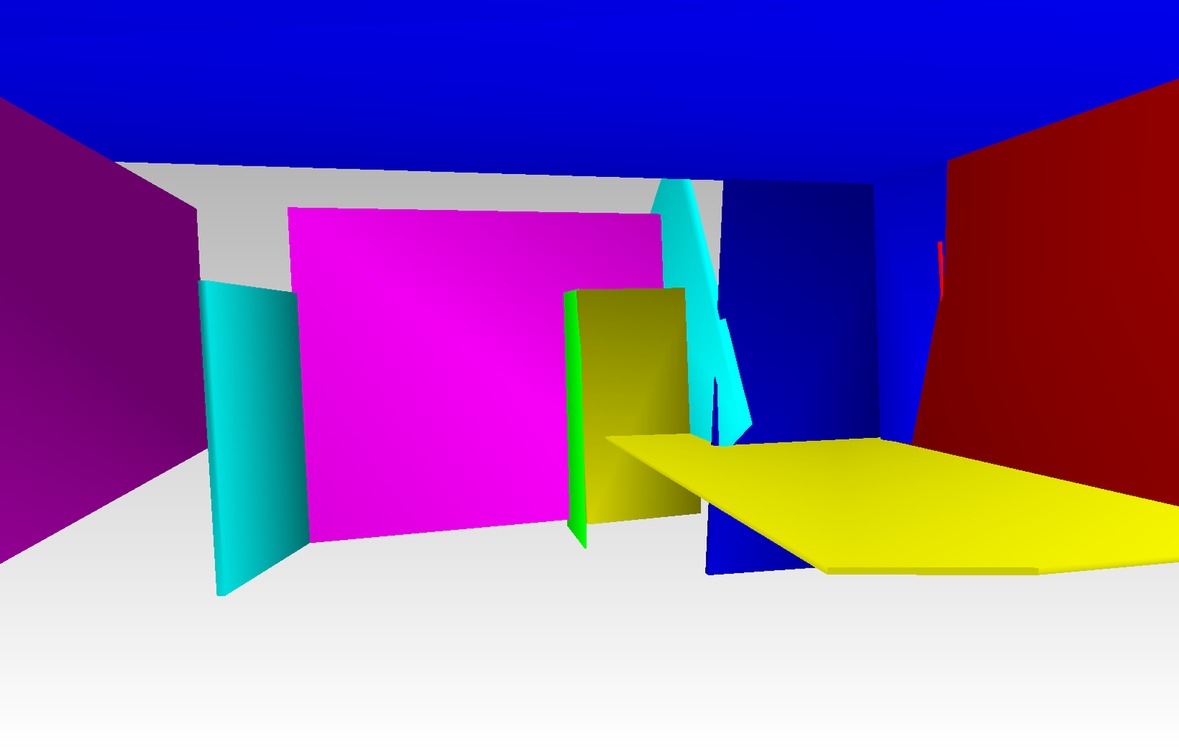}
\end{tabular}
\end{center}
\vspace{-4mm}
\caption{Top row: 3 consecutive views of the Office-P19 dataset, with detected line segments.  Bottom row: views of this area with reconstructed structures: points (left), lines (middle), bounding boxes of planes associated to clusters of coplanar lines after BA (right).}\label{fig::coplanar}
\vspace{-1mm}
\end{figure*}

Another issue concerns the lack of texture in environments such as building interiors, as it greatly reduces the amount of feature points detected in images, also leading to uneven feature distributions.  Besides, if the number of images is small, it is likely that the baselines are wide, as well as the view angles between overlapping images.  Consequently, because of perspective distortion, point descriptors are harder to match; relaxing matching thresholds does not fix the problem, as it introduces outliers.  Moreover, detected points may unknowingly lie on a single plane, possibly giving rise to degenerate configurations for camera registration.  On the contrary, lines do not suffer from lack of texture and are prevalent in interiors, where they often occur at plane intersections and object boundaries, see Fig.~\ref{fig::coplanar}.  Furthermore, lines are robust to significant changes of viewpoints, although their matching eventually also degrades.

Our approach thus leverages on lines for pose estimation and reconstruction, although it also exploits points when available.  Our main contributions are as follow:
\begin{itemize}[itemsep=-3pt,topsep=1pt]
\item Given two successive bifocal calibrations, i.e., sharing a middle camera, we propose a novel method for computing their relative scale without the need for trifocal features. It is based on line coplanarity hypotheses, which is relevant in man-made scenes (cf.\ Sect.~\ref{seccoplanar}).

\item To however exploit trifocality when present, we show how to relax usual trifocal constraints, and integrate them with coplanarity constraints in a unified framework. These relaxed constraints only need one feature triplet and 2 of the bifocal calibrations  
(cf.\ Sect.~\ref{sectrifocal}).

\item Robustness to outliers in this context (cf.\ Sect.~\ref{secrobustestim}) requires parameters that may be hard to set given the variety of scenes and constraint types. We propose a parameterless method that integrates all constraints and adapts automatically to the scene diversity (cf.\ Sect.~\ref{secacrobustestim}).


\item 
Using both standard SfM datasets with ground truth and difficult interior scenes with little overlap and little texture, we validate empirically our technical choices and we show (1) that we can calibrate images that cannot be calibrated with existing point- or line-based SfM methods, and (2) that our accuracy is on par with state-of-the-art SfM methods on less hard datasets.
%
\end{itemize}
Our method does \emph{not} assume any Manhattan configuration.


\section{Related Work}

Incremental SfM methods \cite{Snavely2006Bundler,ACPOINTS,Wu2013linsfm} register a new image to a partial model already constructed, using 3D-2D correspondences.  A number of Perspective-n-Points (PnP) algorithms \cite{Lepetit2008ijcv,Garro20123DV} have been proposed for solving this resection problem\rlap.\,  Whatever the method, at least 3 correspondences common to 3 images are required: three 3D points visible in the same 3 images, and up to minimum of 6 \cite{Zheng2013iccv}.

Hierarchical SfM methods, that additionally merge partial models, also have similar constraints.  In \cite{Havlena2010eccv}, the two models to merge are overlapping in the sense that they share one or several images, and pairs of 3D points projecting to the same 2D features in both models are used to relate the models, which implies tracks of length~3 or more.  In \cite{Toldo2015cviu}, models are merged via feature matches between images separately associated to each model, requiring that the same four 3D points are reconstructed in both models, which implies in turn tracks of length 4 (connections between 2 tracks of length at least 2).  Even with relaxed requirements where merging uses 4 points that are seen but not necessarily reconstructed in the other model \cite{Fusiello2015npnp}, feature tracks of length 3 are required.

As for global SfM methods, their main objective is merging relative motions between two cameras into a consistent graph of all cameras.  Besides robustness concerns, to get rid of outlier edges, and various approaches to average rotations, one of the main issues is that the relative translations are only given up to an unknown scale factor; only their directions are known.  Most methods to infer global translations rely on information redundancy assuming a densely-connected graph \cite{Govindu2001cvpr,Cui2015bmvc}, or on additional information from trifocal tensors \cite{Sim2006cvpr,moulon2013global} (hence requiring 4 tracked points across 3 views).  A number of other methods \cite{KahlH2008pami,Rodriguez2011cvpr,ArieNachimson2012global} compute the global translations, possibly along with the 3D points, by solving equations relating points visible in two images; however, they implicitly assume that enough points are visible in at least 3 images to cancel the degrees of freedom of the relative scale factors.  Besides, they do not all address point match outliers.

The situation is similar with line-based SfM. In \cite{ZHANG_SfM_IJCV14}, an initial image triplet with common line matches is required.  Then, given a partial model, Perspective-n-Line (PnL) methods estimate the pose of a camera in which three 3D lines reproject \cite{Mirzaei2011pnl,Zhang2012accv}, which implies that at least 3 lines are visible in 3 views.  A minimum of 6 lines is sometimes even desirable to prevent noise sensitivity~\cite{Mirzaei2011pnl}, if not~9 for applicability and 25 for accuracy~\cite{Pribyl2015bmvc}.

More generally, when associating both points and lines in a ``Perspective-n-Features'' framework, a minimum of 3 features visible in 3 views is still required \cite{Ramalingam2011icra,Xu2016pnl}.

Our approach for relating the scale of two bifocal calibrations is based on coplanar line pairs.  To our knowledge, coplanar lines have been used for pose estimation, but only in a two-view context and with a Manhattan-world assumption, to identify planar structures \cite{Kim2014accv}.  A related topic is plane-based SfM, but it has mostly been studied assuming prior knowledge (user-given) about the scene planes \cite{Sturm2000cvpr,Bartoli2003ijcv} or in tracking scenarios with videos \cite{Zhou2012cvpr}.  In \cite{Rother2003iccv,KahlH2008pami}, a reference plane is used to estimate both the global translations and 3D points, but it must be visible in all images.

A related work regarding the estimation of scale factors and the identification of planar structures concerns direct structure estimation (DES) via homography estimations, with the computation of coplanar point clusters, but it does not estimate poses and it also relies on trifocal points \cite{Jiang2015cvpr}.

Line triangulation and line bundle adjustment has been well studied given an initial global pose estimation \cite{Bartoli2005}, but not associated to coplanarity issues.

\section{From Relative to Global Pose Estimation}\label{secreltoglob}



We consider the case where the epipolar graph mainly contains long chains or long cycles with little or no trifocal relations.  For every edge in the graph between camera $i$ and camera~$j$, we assume the relative pose $(R_{ij},\td_{ij})$ known (estimated), where $R_{ij}$ is the relative rotation and $\td_{ij}$ is the unit norm translation direction.
We are interested in estimating the scale factors $\la_{ij}$ relating the translation directions $\td_{ij}$ to the global relative translations $\tr_{ij} = \la_{ij} \td_{ij}$, that is, up to a single global scale factor.

In the following, we only consider the case of a single chain or cycle of cameras, for which we estimate global poses $(R_j,\tr_j)$, where rotations $R_j$, translations $\tr_j$ as well as camera centers $C_j$ are defined in the same reference frame.  Yet, our pose estimation method can be integrated in a general global SfM framework for arbitrary graphs, e.g., as described in \cite{Sharp01towardmultiview} to evenly distribute errors over the whole graph in trying to satisfy the coherency constraints:
\begin{align}\label{eqrelglob}
R_j &= R_{ij} R_i, & \tr_j &= R_{ij} \tr_i + \tr_{ij}, & C_j &= -R^\trsp_j \tr_j.
\end{align}
This could be associated to a method to remove outlier edges and enforce cycle consistency \cite{Govindu2006robustness,ZachK2010cvpr,Enqvist2011iccvw}.

When considering a single chain or cycle of cameras, global motions are recursively defined as:
\begin{align}
	R_1 &= I & \text{and}& & R_{j+1} &= R_{j,j+1} R_j \\
  \tr_1 &= 0 & \text{and}& & \tr_{j+1} &= R_{j,j+1}\tr_j + \la_{j,j+1} \td_{j,j+1}
\end{align}
As the global pose remains defined up to an unknown scale factor, we additionally set $\la_{12} = 1$: distances are thus defined with unit length $\la_{12}$. (In case of a cycle, we could also include epipolar constraints to close the loop and distribute errors as in \cite{Sharp01towardmultiview}, but we do not in our implementation.)
Finally, a bundle adjustment refine the initial pose estimation.


In the following, to simplify notations, we assume that features are normalized, i.e., with camera intrinsic parameter matrices $K_j = I$. We also denote an arbitrary triplet of successive cameras $1,2,3$ rather than $j, j\,{+}\,1, j\,{+}\,2$.


\section{Coplanarity Constraint}\label{seccoplanar}

\begin{figure}[!t]
\begin{center}
	\includegraphics[width=0.75\columnwidth]{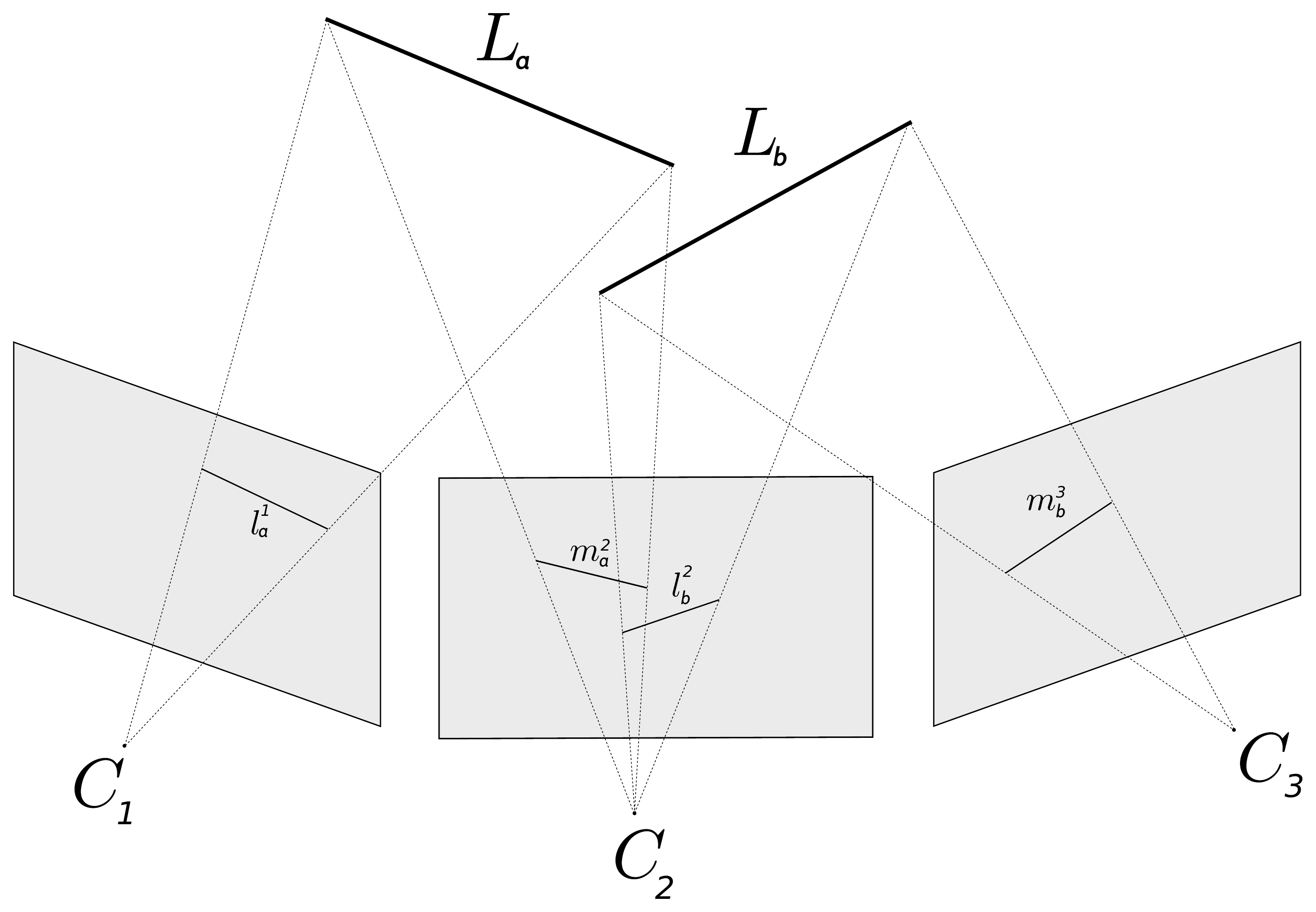}
        \vspace{-2mm}
\end{center}
\caption{If $L_a$ and $L_b$, without trifocal overlap, are known to be coplanar, it is enough to place $C_1,C_2,C_3$ in the same reference frame knowing only bifocal calibrations for cameras 1-2 and 2-3.}
\label{fig::intro}
\end{figure}

Let $L_a$ and $L_b$ be two non-parallel 3D line segments in a plane~$\p$. (Coplanarity here is actually just an hypothesis, to be validated in a RANSAC-like manner, see Sect.~\ref{secrobustestim}.)
Suppose $L_a$ only appears on cameras $1$ and $2$, whereas $L_b$ only appears on cameras 2 and 3. Let $l_i^j$ be the projection of $L_i$ on camera $j$. (See Fig.~\ref{fig::intro}.)

The coordinates of a 3D point $P$ in the global reference frame relates to its projection $p^j$ on camera $j$ and depth $z^j_p$:
\begin{equation}\label{eqproj}
	P = R_j^\trsp(z_p^j p^j - \tr_j)
\end{equation}
Assuming $P\in L_i$, we also have for any camera $k$:
\begin{equation}\label{eqpointinline}
	l_i^{k}\cdot(R_k P + \tr_k) = 0,
\end{equation}
where ``$\cdot$'' denotes scalar product.
Combining~\eqref{eqproj} and \eqref{eqpointinline}, and relating global to relative motions using~\eqref{eqrelglob}, we can express the depth $z^j_p$ in terms of the relative pose information:
\begin{gather}
  l_i^{k}\cdot(R_k R_j^\trsp(z_p^j p^j - \tr_j) + \tr_k) = l_i^{k}\cdot(R_{jk}z_p^j p^j + \tr_{jk}) = 0 \nonumber\\
  \llap{$\Leftrightarrow \quad$} z_p^j = -\frac{l_i^{k}\cdot\tr_{jk}}{l_i^{k}\cdot(R_{jk} p^j)}.\label{eqz}
\end{gather}
%
%
We also know that the normal to plane~$\p$ is given by: 
\begin{equation}
	\np = d_{L_a} \times d_{L_b} = (R_1^\trsp l_a^1 \times R_2^\trsp l_a^2) \times (R_2^\trsp l_b^2 \times R_3^\trsp l_b^3)
\end{equation}
where $d_{L_i}$ is the 3D direction of line $L_i$.  Let $M$ be a point in~$\p$.  As any line in $\p$ is orthogonal to $\np$, we have:
\begin{equation}
	P \in \p \Leftrightarrow \np\cdot(P-M) = 0
\end{equation}
Since $P \in L_i \subset \p$, using~\eqref{eqproj} and~\eqref{eqz} yields:
\begin{gather}
\np\cdot(R_j^\trsp(z_p^j p^j - \tr_j)-M) = 0 \rlap{$\quad\Leftrightarrow$}
\nonumber\\
\np\cdot(-\frac{l_i^{k}\cdot\tr_{jk}}{l_i^{k}\cdot(R_{jk} p^j)}   R_j^\trsp p^j + C_j - M) = 0 \rlap{$\quad\Leftrightarrow$}
\nonumber\\\label{eqnpcjs}
\np\cdot(C_j - M) = \frac{(l_i^{k}\cdot\tr_{jk})(\np\cdot (R_j^\trsp p^j))}{l_i^{k}\cdot(R_{jk} p^j)}
\end{gather}
Now using~\eqref{eqnpcjs} both for a point $P_a \,{\in}\, L_a$ with $j\,{=}\,2$ and $k\,{=}\,1$, and a point $P_b \,{\in}\, L_b$ with $j\,{=}\,2$ and $k\,{=}\,3$, we get:
\begin{gather}
\frac{(l_b^3\cdot\tr_{23})(\np\cdot(R_2^\trsp p_b^2))}{l_b^3\cdot(R_{23} p_b^2)} = \frac{(l_a^1\cdot\tr_{21})(\np\cdot(R_2^\trsp p_a^2))}{l_a^1\cdot(R_{21} p_a^2)}
\rlap{$\quad\Leftrightarrow$} \nonumber\\
\frac{l_b^3\cdot\tr_{23}}{l_a^1\cdot\tr_{21}} = \frac{(l_b^3\cdot(R_{23} p_b^2))(\np\cdot(R_2^\trsp p_a^2))}{(l_a^1\cdot(R_{21} p_a^2))(\np\cdot(R_2^\trsp p_b^2))}
\rlap{$\quad\Leftrightarrow$} \nonumber\\ \label{eqratiolambda}
\frac{\la_{23}}{\la_{21}} = \frac{(l_b^3\cdot(R_{23} p_b^2))(\np\cdot(R_2^\trsp p_a^2))(l_a^1\cdot\td_{21})}{(l_a^1\cdot(R_{21} p_a^2))(\np\cdot(R_2^\trsp p_b^2))(l_b^3\cdot\td_{23})}
\end{gather}
Eq.~\eqref{eqratiolambda} expresses the signed ratio of the distances between 3 successive camera centers in the same reference frame, in terms of only 2-view relative pose information, plus weak information about 2 coplanar lines that do not have to be visible in all 3 views.
It has 3 degenerate configurations that can be assessed and controlled with an angular threshold.

Note that this criterion does not depend on the endpoints of line segments, that are notoriously inaccurate.  Line segment matches can actually be grossly wrong because of over-segmentation, which is a common weakness of line segment detectors.  On the contrary, our criterion only relies on (infinite) lines, which are much more stable.  It can cope with line segments matches that are wrong w.r.t.\ endpoints but correct w.r.t.\ the supporting 3D line.


\section{Simplified Trifocal Constraints, When Any}\label{sectrifocal}

In case a line or point is visible in three successive views 1-2-3, we consider a corresponding simplified trifocal constraint, that also involves the scale factors $\la_{12},\la_{23}$. For both kinds of features, we do as follows:
\begin{enumerate}[itemsep=-2pt,topsep=1pt]
\item We set arbitrarily the first two camera centers given their relative pose: $C_1 = 0$, $\la_{12}=1$, $C_2 = -R_2^T \tr_{12}$.
\item We compute the approximate 3D reconstruction of the feature from the first 2 camera poses, as well as its projection on the third camera.
\item We compute the scale factor $\la_{23}$ such that the projected feature on camera 3 corresponds best to the detection.
\end{enumerate}
%
%
As in step 1 we set $\la_{12} = 1$, the computed scale factor $\la_{23}$ is actually also the ratio $\tau_{123} = \la_{23}/\la_{12}$. For a more robust and accurate estimation, we can symmetrize the setting: permuting the role of cameras and averaging the resulting scale factors.  As we just consider here a chain of successive views, it makes sense to actually only take into account the case where cameras 1 and~3 are swapped, keeping camera~2 as central. It prevents camera configurations 1-3 or 3-1, which are likely to feature wider changes of viewpoints, i.e., wider baselines and view angles.  Bifocal calibration 1-3 may even fail and not be available.  (In our implementation, we thus only compute the ratios $\tau_{123}$ and $\tau_{321}$, retaining the average $\tau = (\tau_{123} + 1/\tau_{321})/2$.)
Note that, as opposed to usual trifocal constraints that require at least three trifocal features, one is enough here.

In step 3, as detailed below, the scale factor $\la_{23}$ is expressed in the form:
\begin{equation}\label{eqargmin}
	\la_{23} 
	= \argmin_{\la \in \mathbb{R}} \frac{\| u \times (v + \la w) \|}{\| u \| \, \|v + \la w \|},
\end{equation}
where $u,v,w$ are known vectors in $\mathbb{R}^3$. To find the minimum, we look for the values of $\la$ that make the derivative vanish. It leads to a third-degree polynomial, that simplifies into a second-degree polynomial, whose roots can be tested to find the minimum of the original expression.
As above, this formula has a degenerate configuration that can be checked and discarded using an angular threshold.

\smallsection{Trifocal point constraint}\label{sectrifpoint}
%
%
Given a triplet of matched points $\hp = (p_1,p_2,p_3)$ in cameras 1-2-3, we can estimate the corresponding 3D point $\tP$ from $p_1,p_2$, assuming $\la_{12} = 1$, and reproject $\tP$ as $\tp_3$ on camera 3, for some given $\la_{23}$:
\begin{equation*}
	\tp_3 = R_3(\tP-C_3)  = R_3(\tP - C_2) - \la_{23} \td_{23}
\end{equation*}
We look for the scale factor $\la_{23}^*$ that makes $\tp_3$ the closest to the observation $p_3$.  Rather than minimizing the distance between the two 2D points, we minimize the angle $\widehat{p_3 C_3 \tp_3}$:
\begin{align}
	\la_{23}^* 
	&= \argmin_{\la_{23} \in \mathbb{R}} \frac{\| p_3 \times \tp_3 \|}{\| p_3 \| \, \|\tp_3 \|} \nonumber\\
	&= \argmin_{\la_{23} \in \mathbb{R}} \frac{\| p_3 \times (R_3(\tP - C_2) - \la_{23} \td_{23}) \|}{\|p_3\| \, \|R_3(\tP - C_2) - \la_{23} \td_{23}\|},
\end{align}
which is in the form~\eqref{eqargmin}.

\smallsection{Trifocal line constraint}\label{sectrifline}
%
%
Given a triplet of matched line segments $\hl = (l_1,l_2,l_3)$ in cameras 1-2-3, we can estimate the corresponding 3D line $\tL$ from $l_1,l_2$, assuming $\la_{12} = 1$, and reproject $\tL$ as $\tl_3$ on camera 3, for some given $\la_{23}$. As $\tl_3$ also is the normal of the plane defined by $C_3$ and $\tL$, we have for any point $\tP$ on $\tL$:
\begin{equation*}
	\tl_3 \propto R_3 [d_{\tL} \times (\tP - C_3)] 
\end{equation*}
We look for the scale factor $\la_{23}^*$ that makes $\tl_3$ the most colinear with the observation $l_3$, i.e., that minimizes the relative angle between $l_3$ and $\tl_3$:
\begin{align}
	\la_{23}^*
	&= \argmin_{\la_{23} \in \mathbb{R}} \frac{\| l_3 \times \tl_3 \|}{\| l_3 \| \, \|\tl_3 \|} \nonumber\\
	&= \argmin_{\la_{23} \in \mathbb{R}} \frac{\| l_3 \times (R_3 [d_{\tL} \times (\tP - C_3)]) \|}{\|l_3\| \, \|R_3[d_{\tL} \times (\tP - C_3)]\|} \nonumber\\
	&= \argmin_{\la_{23} \in \mathbb{R}} \frac{\| l_3 \times (R_3 [d_{\tL} {\times} (\tP {-} C_2)] - \la_{23} R_3 [d_{\tL} {\times} \td_{23}]) \|}{\|l_3\| \, \|R_3 [d_{\tL} {\times} (\tP {-} C_2)] - \la_{23} R_3 [d_{\tL} {\times} \td_{23}]\|}
\end{align}
\vskip-0.7\baselineskip\noindent
which is in the form of~\eqref{eqargmin}.
%
%
As for coplanarity (Sect.~\ref{seccoplanar}), this constraint does not depend on line segment endpoints.


\section{Robust Estimation}\label{secrobustestim}

As there is no oracle to safely pick non-parallel coplanar lines (cf.\ Sect.~\ref{seccoplanar}), we adopt a RANSAC-like approach to sample candidate line pairs and select the associated scale $\la_{23}$ with which the largest number of pairs agree.  More generally, we also sample and check agreement w.r.t.\ trifocal points and lines when any (cf.\ Sect.~\ref{sectrifocal}), which provides robustness to wrong detections and matching too.  For this, we have to define a measure of residual error for the three different kinds of features (hypothesized coplanar line pairs, trifocal points, trifocal lines), given a presumed $\la_{23}=\la$ obtained from the sample (discarding degenerate cases).

\smallsection{Coplanar lines}\label{secerrorcoplanar}
For any quadruplet of line segments $\cl = (l_a^1,l_a^2,l_b^2,l_b^3)$ s.t. $l_a^1,l_a^2$ match in cameras 1-2 and $l_b^2,l_b^3$ match in cameras 2-3, we estimate the associated 3D lines $L_a,L_b$ and consider the 3D point $P_{ab}{\,\in\,} L_a$ (resp.\ $P_{ba}{\,\in\,} L_b$) that is the closest to $L_b$ (resp.\ $L_a$) in 3D, with $P_{ab}{=}P_{ba}$ if $L_a,L_b$ are coplanar. We then consider their reprojection $\tp_{ab}^2,\tp_{ba}^2$ on the shared camera 2. The residual error is their pixel distance: $d\co(\cl) = d\co(L_a,L_b) = d(\tp_{ab}^2,\tp_{ba}^2)$.

Considering $\tp_{ab}^2,\tp_{ba}^2$ rather than $P_{ab},P_{ba}$ removes one dimension of error, along the direction of~$C_2$, possibly constraining $\lambda$ less. But positioning by triangulation along this direction is generally less accurate and thus less meaningful, leading people to rather use reprojected distances and rely on other views to capture any error along this direction.


To avoid degenerate cases with mostly parallel lines, we also discard line pairs whose 3D directions are similar.  (In our experiments, we use a threshold of $15^\circ$.)  Note that these directions can be computed from the global rotations only, before global translations are estimated.  As there can be many pairs candidate for coplanarity, we only consider, for each segment in camera~2, having a match in camera~$j$, its $N$ closest neighbors having a match in the other camera, with a distance defined as the minimum distance between segment endpoints. (In our experiments, $N{=}10$).

\smallsection{Trifocal point error}\label{secerrorpoint}
For each triplet of matched points $\hp = (p_1,p_2,p_3)$ in cameras 1-2-3, we estimate the corresponding 3D point $P$ from $p_1,p_2$, and reproject it as $\tp_3$ on camera 3 (assuming $\la_{23}$), as in Sect.~\ref{sectrifpoint}.  The residual error is the pixel distance of reprojection $\tp_3$ to the observed detection $p_3$.  For robustness, we actually symmetrize this measure by swapping cameras 1 and 3, and averaging the distances: $d\pt(\hp) = (d(\tp_1,p_1) + d(\tp_3,p_3))/2$.

\smallsection{Trifocal line error}\label{secerrorline}
For each triplet of matched line segments $\hl = (l_1,l_2,l_3)$ in cameras 1-2-3, we estimate the corresponding 3D line $L$ from $l_1,l_2$, and reproject it as $\tl_3$ on camera 3 (assuming $\la_{23}$), as in Sect.~\ref{sectrifline}. The residual error is the average of the pixelic distance between the reprojected (infinite) line $\tl_3$ and the two endpoints of the detected segment $l_3$, as in \cite{Bartoli2005,ZHANG_SfM_IJCV14}.  For robustness, we also swap cameras 1 and 3, and define $d\se(\hl)$ as the average of both errors.

\smallsection{Sampling}\label{secsampling}
In practice, the total number of features (hypothesized coplanar line pairs, trifocal points, trifocal lines) is generally low enough (usually less than 10,000) for all features to be tried rather than sampled.  


\section{Parameter-Free Robust Estimation}\label{secacrobustestim}

Rather than depend on fixed arbitrary error thresholds to define feature agreement with a model, we actually resort to an \emph{a contrario} (AC) approach \cite{ACRANSAC,ACPOINTS}: we compute the expectation of the number of false alarms (NFA), that measures the statistical meaningfulness (actually the converse, i.e., the insignificance) of $\la_{23}$ w.r.t.\ to the features to test, and select the scale $\la_{23}$ with the lowest NFA. It allows an automatic optimization of the inlier-outlier threshold, and thus more accurate results too.  This is also consistent with the method we use for two-view pose estimation \cite{ECCV16}.



\textbf{Coplanar line NFA. }
For coplanarity, we follow \cite{ECCV16} and define a line error from the error for line pairs:
\begin{equation}
	d\co(L_i, \la) = \min_{L_j \text{ coplanar with } L_i} d\co(L_i, L_j, \la)
\end{equation}
where $d\co(L_i, L_j, \la)$ is the coplanarity distance defined in Sect.~\ref{secerrorcoplanar}.  The NFA for our sampling is then, following \cite{ACPOINTS}:
\begin{equation}
	\!\! \NFA\co(\la) {\,=\,} (n\setwo{-}2) \min_{\!\!k \in [3, n\co]\!\!} n\setwo N \dbinom{n\setwo}{k{-}2} \!\!\left[\frac{\pi \, d\co(L_k, \la)^2}{\mathcal{A}}\right]^{k-2}
\end{equation}
where $n\setwo$ is the number of lines in camera~2 that have at least a match in camera 1 or~3, $\mathcal{A}$ is the image area, and $L_k$ is the $k$-th best inlier (with lowest error).



\textbf{Trifocal point NFA. }
For triplets of matched points, following \cite{ACPOINTS}, we have:
\begin{equation}
	\NFA\pt(\la) = (n\pt{-}1) \min_{\!\!k \in [2, n\pt]\!\!} \dbinom{n\pt}{k} k \left[\frac{\pi \, d\pt(\hp_k, \la)^2}{\mathcal{A}}\right]^{k-1}
\end{equation}
where $n\pt$ is the total number of matched point triplets in cameras 1-2-3, $\mathcal{A}$ is the image area, $d\pt(\hp, \la)$ is the residual error of triplet $\hp$ assuming scale $\la$ (cf.\ Sect.~\ref{secerrorpoint}), and $\hp_k$ is the $k$-th lowest error.


\textbf{Trifocal line NFA. }
For triplets of matched line segments, also following \cite{ACPOINTS}, we have:
\begin{equation}
	\NFA\se(\la) = (n\se{-}1) \min_{\!\!k \in [2, n\se]\!\!} \dbinom{\!n\se\!\!}{k} k \!\left[\!\frac{2\mathcal{D} \, d\se(\hl_k, \la)}{\mathcal{A}}\!\right]^{k-1}
\end{equation}
where $n\se$ is the total number of matched line triplets, $\mathcal{D}$ is the diagonal length of the picture, $d\se(\hl, \la)$ is the residual error of triplet $\hl$ assuming scale $\la$ (cf.\ Sect.~\ref{secerrorline}), and $\hl_k$ is the $k$-th lowest error.

\smallsection{Global NFA}
As these NFAs correspond to expectations of independent events, the global NFA is their product:
\begin{equation}
	\NFA(\la) = \NFA\co(\la) . \NFA\pt(\la) . \NFA\se(\la).
\end{equation}
It estimates the insignificance of a candidate ratio~$\la$ based on coplanarity and trifocal constraints, when any.  For robust estimation, we retain the $\la$ with the overall lowest NFA. It defines a parameterless AC-RANSAC variant to Sect.~\ref{secrobustestim}.

Note that this pose estimation method can be heavily parallelized, not only to evaluate the different constraints but also to estimate the global motion for any three consecutive views, to be later combined in a chain as defined in Sect.~\ref{secreltoglob}.

\section{Bundle Adjustment}\label{bundleadjustement}

The last step of our SfM method is a bundle adjustment (BA) that refines simultaneously the structure and the poses.

\smallsection{Reconstructed structure}
In our context, we could consider as structures not only points and lines, but also planes corresponding to coplanar lines.  Indeed, just as tracks of points or lines across images represent single structures, single planes could be associated to sets of lines sharing a coplanarity constraint.  More precisely, only coplanar line pairs with similar plane orientations should be clustered together, as a line can belong to two different planes at edges, e.g., where a wall meets another wall, floor or ceiling.

Yet, we observed on real scenes that such a clustering of coplanar lines into single planes tends to degrade the accuracy of pose estimation.  Our interpretation is that individual pairs of (real 3D) lines can be coplanar enough to robustly estimate sensible scale ratios but, when grouped in a single plane, their global coplanarity deteriorates.

In fact, contrary to 3D points, that are well determined although possibly misdetected or mismatched in images, there is no such thing \emph{in the real world} as perfect 3D lines, perfect 3D planes, nor exact line parallelism, orthogonality or coplanarity.  (Optical distortion and detection noise just come on top of it.)  This is especially true of edges and surfaces in a building, as tolerances of straightness and flatness in the construction industry range typically from 0.2 to 1\%.  Line-based calibration is thus prone to be less accurate in practice than point-based calibration, and even less when two lines are involved in a feature, as in line coplanarity.

Buildings also present many cases of near colinearity, hence near line coplanarity, due to close edges at the boundary of thin surfaces, e.g., baseboards, conduits, moldings, picture frames, whiteboards, window and door frames, etc.  Furniture edges also tend to be almost but not exactly coplanar with wall edges.  Due to small errors, such nearly coplanar lines are considered as inliers, but degrade accuracy.

This consideration is consistent with \cite{ECCV16} where the authors observe that, in real data, lines that should ``logically'' be parallel turn out not to be as much as expected, leading to a number of close but different-enough vanishing points (VPs).  They found however that treating parallel line pairs independently leads to more accurate calibrations than merging them into a single VP.
Similarly, in our bundle adjustment, we do not consider planes as the support of many coplanar lines; line pairs that were determined as coplanar, i.e., RANSAC inliers, are treated individually. 

\smallsection{Residuals and optimization}
The parameters of our BA are tracked points and lines, as well as camera positions and orientations.  The error to minimize is the sum of the square (pixelic) distance of the reprojected points and lines to their detection (cf.\ Sect.~\ref{secerrorcoplanar}), in all the cameras that see (i.e., detect and match) them, plus the sum of the square coplanarity residuals $d\co(L_a,L_b)$ (cf.\ Sect.~\ref{secerrorcoplanar}), for all line pairs $L_a,L_b$ found as inliers, in all images that see both lines.

Concretely, we initialize the bundle adjustment with the pose found by composing the scaled relative motions (cf.\ Sect.~\ref{secreltoglob}) and with triangulated features. As BA can be sensitive to initialization and given that rotations are often better estimated than translations, we first refine the structure and motion with fixed rotations, then refine all parameters (as in \cite{Olsson2011scia,moulon2013global}). We use the Ceres solver \cite{ceres} for minimization.


\section{Experiments}\label{secexp}


\smallsection{Feature detection and matching}
Points are detected and matched with SIFT \cite{SIFT}. Lines are detected with MLSD \cite{multiLSD} and matched with LBD \cite{LBD}.  Both kinds of features are tracked across consecutive pictures to identify match triplets (hence trifocal overlaps), when any.
The code is available on GitHub \footnote{(\url{https://github.com/ySalaun/LineSfM})}.


\smallsection{Bifocal calibration}
We implemented our SfM approach on top of the two-view relative pose estimation of \cite{ECCV16}.  Besides being parameterless, it has the advantage of robustly combining both line and point features, when any, providing state-of-the-art accuracy even in textureless environments and with wide baselines.  When points are not available, it only assumes that both images contains at least two pairs of matched lines that are parallel in 3D; this constraint is most often met in indoor scenes.  If not met, our assumption (seeing lines in views 1-2 coplanar with lines seen in views 2-3) is likely not to be met either.

Using method \cite{ECCV16} is not intrinsic to our approach.  We could have used just as well any other bifocal calibration method, e.g., based on points assuming enough (5 or 7) are available on all image pairs, or based on lines as in \cite{CVPR11}, although it assumes a Manhattan scene, contrary to \cite{ECCV16}.

\begin{table*}[!t]
\begin{center}
\begin{tabular}{c@{~~~~}c}
\begin{tabular}{|c|@{~}c@{~}|@{~}c@{~}|@{~}c@{~}|@{~}c@{~}|@{~}c@{~}|@{~}c@{~}|}
\hline
\multirow{2}{*}{\backslashbox[22mm]{Scene}{Method}} & \multicolumn{5}{@{~}c@{~}|}{RANSAC threshold (pixels)} & AC-RAN \\
\cline{2-6}
 & 0.5 & 1 & 3 & 6 & 9 & \!\!SAC\!\!\\
\hline
\hline
Castle P19 & 146 & 90.5 & 107 & \textbf{90} & 196 & 101 \\
\hline
Castle P30 & 91 & 83 & 73 & 76 & 144 & \textbf{72}\\
\hline
Entry P10 & \textbf{7.6} & 7.9 & 8.4 & 9.3 & 10.6 & 7.8\\
\hline
Fountain P11 & 2.2 & \textbf{2.1} & 2.4 & 2.5 & 3.8 & 2.3\\
\hline
Herz-Jesu P8 & 4.0 & 4.1 & \textbf{3.9} & 7.1 & 6.4 & 4.2\\
\hline
Herz-Jesu P25 & 8.5 & 9.0 & 8.6 & 13.3 & 16.9 & \textbf{7.9} \\
\hline
\hline
Average & 43.2 & 32.8 & 33.9 & 33.0 & 63.0 & \textbf{32.5}\\
\hline
\end{tabular}
&
\begin{tabular}{|@{~}c@{~}|@{~}c@{~}|@{~}c@{~}|@{~}c@{~}|@{~}c@{~}|@{~}c@{~}||@{~}c@{~}|}
\hline
\multirow{2}{*}{\backslashbox[22mm]{Scene}{Feature}} & \multirow{2}{*}{Points} & \multirow{2}{*}{Lines} & Points & \multirow{2}{*}{Copl.} & \multirow{2}{*}{All} & All \\
& & & \!\!+lines\!\! & & & +BA \\
\hline
\hline
Castle P19 & 97.9 & \textbf{80} & 98 & 129 & 101 & 22.2 \\
\hline
Castle P30 & 80.3 & 111.2      & 78 & 134 & \textbf{72} & 23.8\\
\hline
Entry P10 & 7.9 & 12.1         & 8.0 & 8.9 & \textbf{7.8} & 5.6\\
\hline
Fountain P11 & \textbf{2.3} & 2.5 & \textbf{2.3} & 52 & \textbf{2.3} & 2.4 \\
\hline
Herz-Jesu P8 & \textbf{4.1} & 4.3 & \textbf{4.1} & 5.4 & 4.2 & 3.8 \\
\hline
Herz-Jesu P25 & 8.4 & 15.4 & 8.4 & 36 & \textbf{7.9} & 5.4\\
\hline
\hline
Average & 33.5 & 37.6 & 33.1 & 60.9 & \textbf{32.5} & 10.5 \\
\hline
\end{tabular}
\end{tabular}
\end{center}
\vspace{-2mm}
\caption{Left: comparison of RANSAC with a single fixed threshold to AC-RANSAC. Right: contribution of the different kinds of features. The numbers measure the average position error of cameras (in mm) before bundle adjustment. Best results for given scene in \textbf{bold}.}
\label{tab::ransac_features}
\vspace{-0mm}
\end{table*}

\begin{table*}[!t]
\begin{center}
\begin{tabular}{|c||c|c|c|c|c|c||c|c|}
\hline
\multirow{2}{*}{\backslashbox[25mm]{Scene}{Method}} & Our & \!\!OpenMVG\!\! & \!Olsson\! & Cui & Arie & Jiang & \!Bundler\! & VSfM \\
& method & \cite{moulon2013global} & \cite{Olsson2011scia} & \cite{Cui2015bmvc} & \cite{ArieNachimson2012global} & \cite{Jiang2013iccv} & \cite{Snavely2006Bundler} & \cite{Wu2013linsfm}\\
\hline
Castle P19 & \textbf{22.2} & 25.6 & 76.2 & --- & --- & --- & 344 & 258 \\
\hline
Castle P30 & 23.8 & 21.9 & 66.8 & \textbf{21.2} & --- & 235 & 300 & 522 \\
\hline
Entry P10 & \textbf{5.6} & 5.9 & 6.9 & --- & --- & --- & 55.1 & 63.0\\
\hline
Fountain P11 & 2.4	& 2.5 & \textbf{2.2} & 2.5 & 4.8 & 14 & 7.0 & 7.6\\
\hline
Herz-Jesu P8 & 3.8 & \textbf{3.5} & 3.9 & --- & --- & --- & 16.4 & 19.3\\
\hline
Herz-Jesu P25 & 5.4 & 5.3 & 5.7 & \textbf{5.0} & 7.8 & 64 & 21.5 & 22.4\\
\hline
\end{tabular}
\end{center}
\vspace{-2mm}
\caption{Average position error: comparison with global \cite{moulon2013global,Olsson2011scia,Cui2015bmvc,ArieNachimson2012global,Jiang2013iccv} and incremental \cite{Snavely2006Bundler,Wu2013linsfm} SfM methods. Best results in bold.}
\label{tab::overall}
\vspace{-2mm}
\end{table*}

\begin{figure}[!t]\columnsep0pt
\mbox{}\!\!\begin{tabular}{c@{~}c}
	\includegraphics[width=0.48\columnwidth]{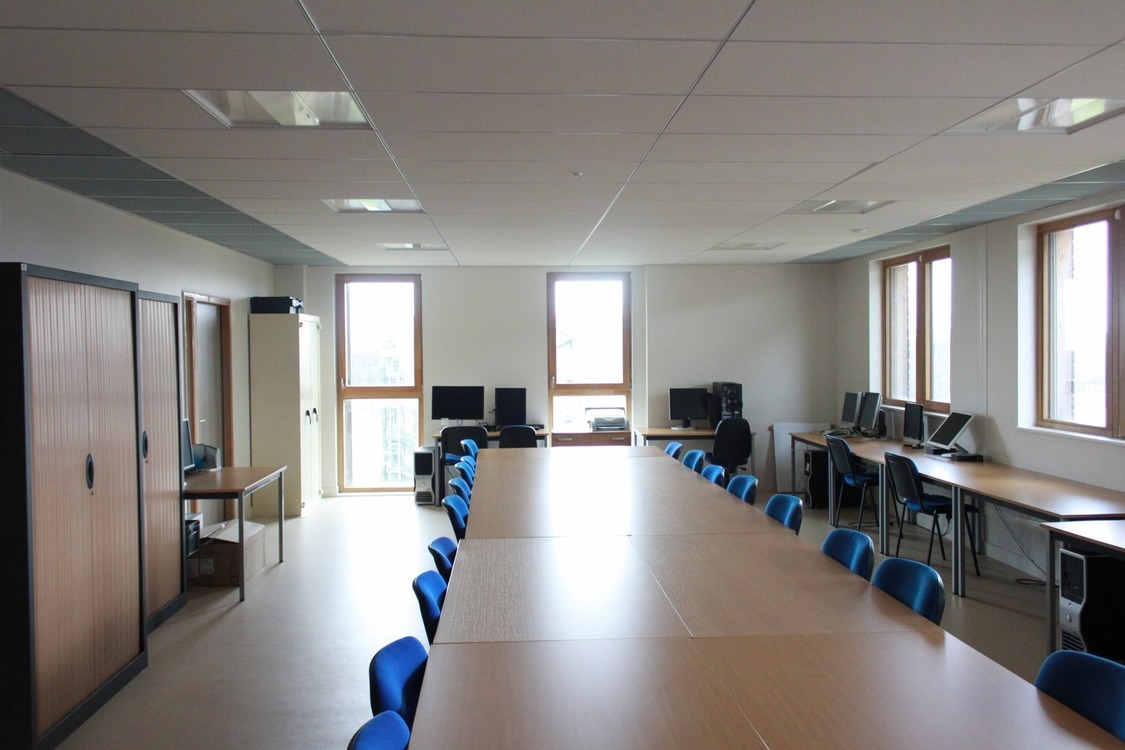} &
	\includegraphics[width=0.48\columnwidth]{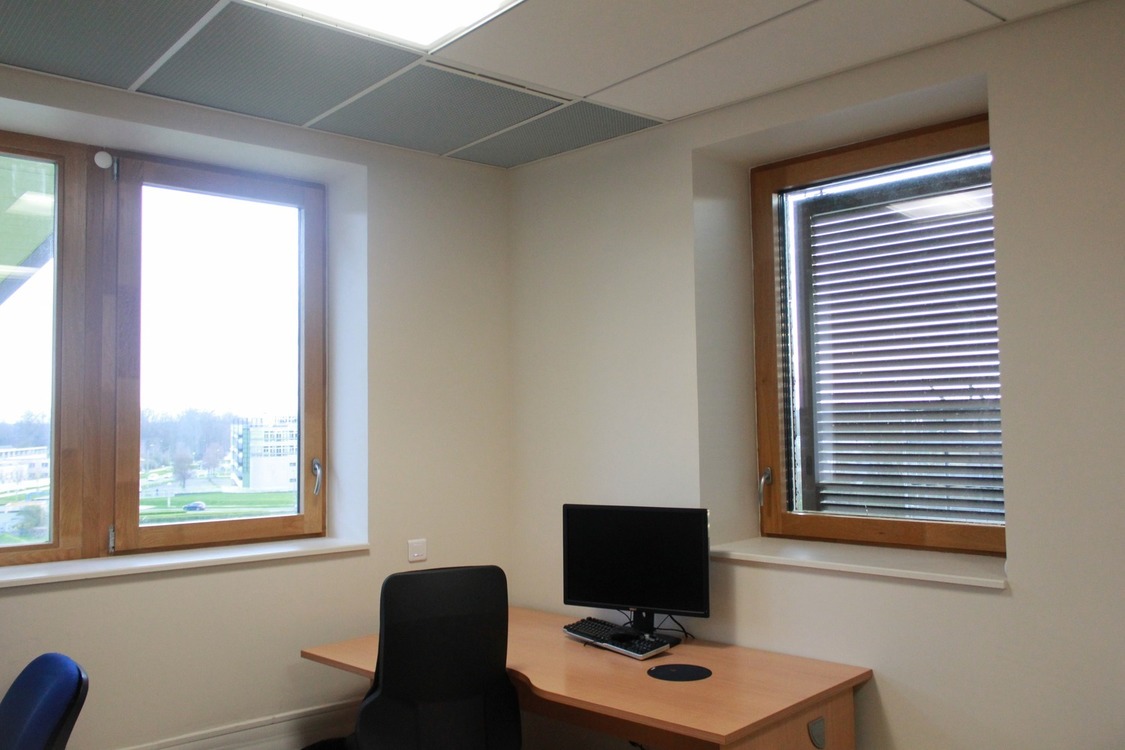} \\
\end{tabular}
\caption{Left: Meeting-P31. Right: Trapezoid-P17 (non-Manhattanness can be seen on the ceiling at room corner).}
\label{fig::indoor}
\vspace{-2mm}
\end{figure}

\smallsection{Datasets}
We consider both difficult interior scenes and a standard SfM dataset of outdoor scenes with ground truth.
\begin{itemize}[nolistsep]
\item The indoor dataset pictures various office rooms with little texture and little image overlap. Office-P19 (cf.\ Fig.~\ref{fig::coplanar}), Meeting-P31 and Trapezoid-P17 (cf.\ Fig.~\ref{fig::indoor}) consist of cycles of images.  P$n$ means $n$ pictures.  Trapezoid-P17 does not belong to a Manhattan world. Resolution is $5184 {\,\times\,} 3456$. As can be seen on Fig.~\ref{fig::coplanar}, points are dense on some textured objects, like the door, but scarce in large other parts of the scene, e.g., white walls.

\item Strecha \etal's dataset \cite{STRECHA} is a de facto standard for assessing the accuracy of SfM. It consists of 6 outdoor scenes with ground truth for camera poses.  We consider both the full dataset as well as subsets of images to reduce image overlap. Resolution is $3072 {\,\times\,} 2048$.
\end{itemize}
All images have been corrected for radial distortion.

\smallsection{RANSAC with and without parameters}
Table~\ref{tab::ransac_features} (left) compares RANSAC with a single fixed threshold for all three kinds of features (cf.\ Sect.~\ref{secrobustestim}) to the parameterless AC-RANSAC (cf.\ Sect.~\ref{secacrobustestim}).  Although AC-RANSAC does not always yield the best results, it works better on average.  For a given scene, a better accuracy could be achieved by setting the different thresholds for each kind of features, but it would not be practical, hence the interest of AC-RANSAC.  In the following, all experiments rely on AC-RANSAC.

\smallsection{Contribution of the different kind of features}
Table~\ref{tab::ransac_features} (right) reports the accuracy of the different kinds of features alone, or when used jointly.  When studying coplanarity features alone, the residual includes the reprojection error of coplanar lines (which may or may not be trifocal).

Trifocal line features provide just a little less accuracy than trifocal points, which are prevalent in this textured datatset.  Used together, they lead to a slightly better accuracy.  Coplanarity features alone yield a little less accuracy than trifocal lines; only Fountain-P11 has a poor calibration with coplanarity, probably because it contains less planes than other scenes.  Yet, after being merged with other features, coplanarity does not degrade accuracy in general, and often even improves it.


\begin{table}[t]
\begin{center}
\begin{tabular}{|c|c|c|}
\hline
\backslashbox[25mm]{Scene}{Method} & Our method & OpenMVG \cite{moulon2013global} \\
\hline
\hline
Castle P19 & 0.30 & 0.21\\
\hline
Castle P30 & 0.29 & 0.21\\
\hline
Entry P10 & 0.006 & 0.90\\
\hline
Fountain P11 & 0.26 & 0.19\\
\hline
Herz-Jesu P8 & 0.28 & 0.20\\
\hline
Herz-Jesu P25 & 0.29 & 0.20\\
\hline
\hline
Office P19 & 1.07 & \color{red}6/20 \\
\hline
Meeting P31 & 1.79 & \color{red}9/31\\
\hline
Trapezoid P17 & 3.60 & \color{red}3/17\\
\hline
\end{tabular}
\end{center}
\caption{Mean reprojection error of all 3D points of all cameras (in pixels).  In {\color{red}red}, fraction of calibrated cameras in case of failure.  Our residual error in indoor scenes is about 4 to 10 times larger than on Strecha's dataset, but picture resolution is 3 times as large.}
\label{tab::residuals}
\vspace{-2mm}
\end{table}

\begin{figure*}
\begin{center}
\begin{tabular}{@{}c@{~~}c@{~~}c@{}}
	\includegraphics[width=0.7\columnwidth]{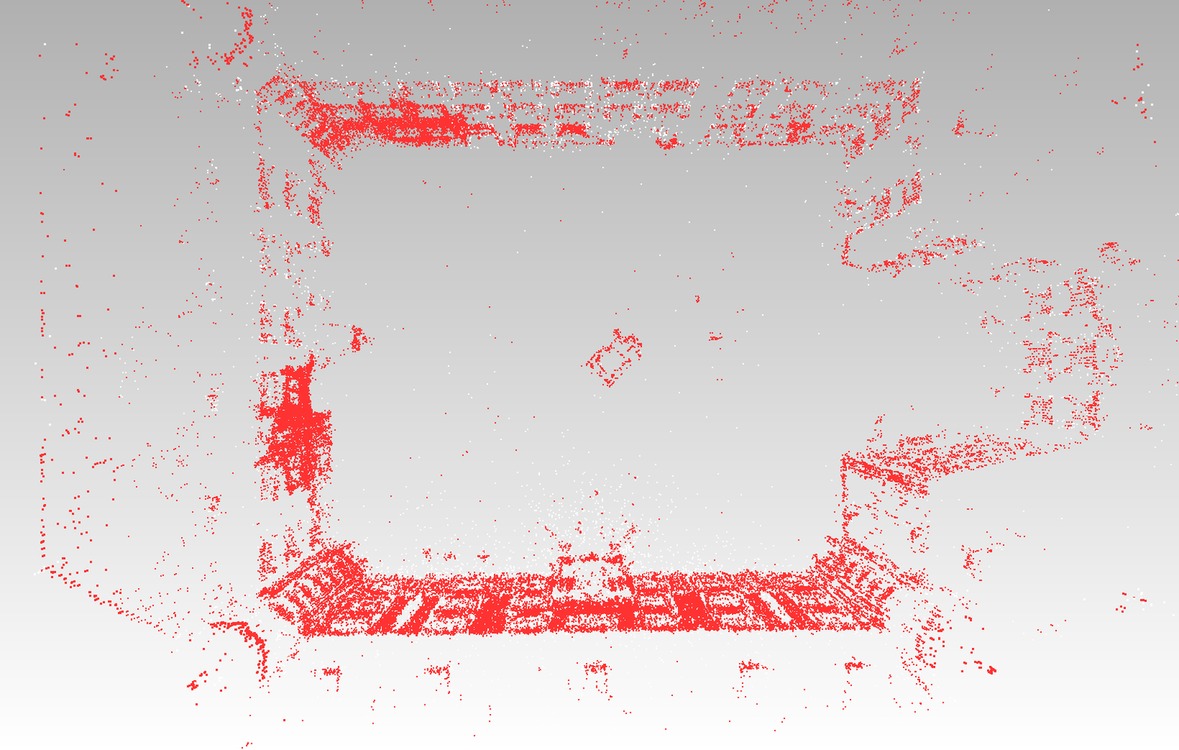} &
	\includegraphics[width=0.7\columnwidth]{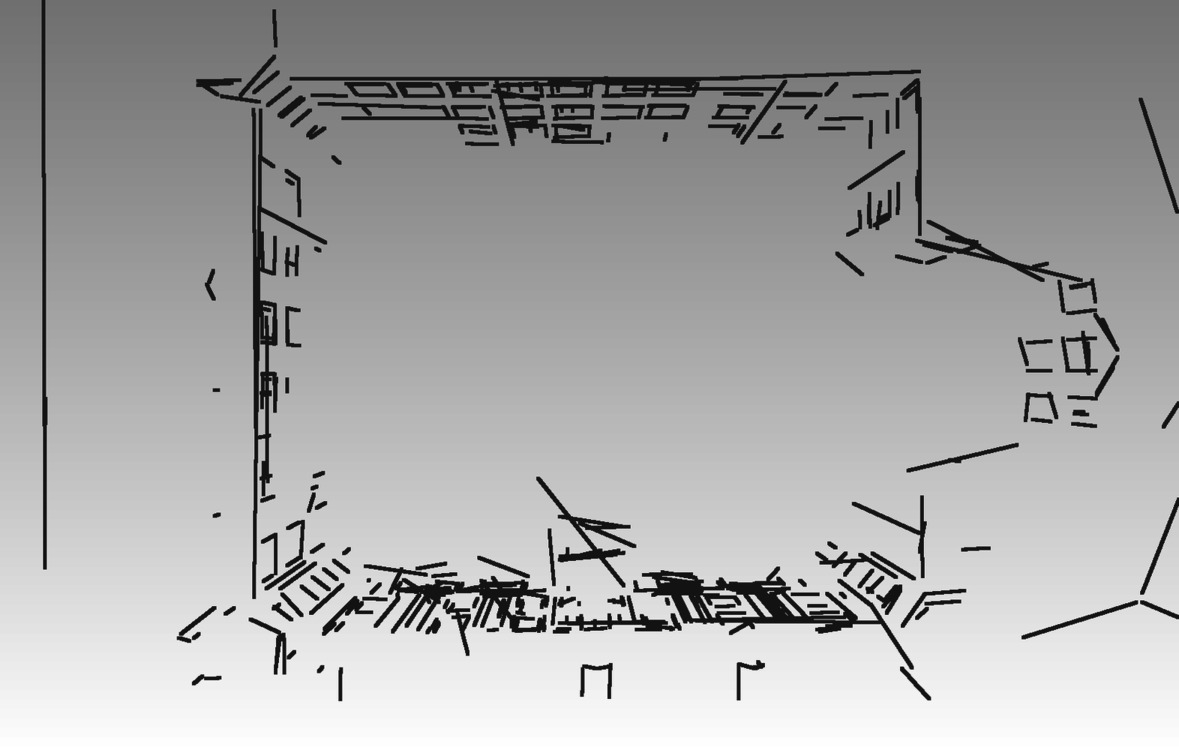} &
	\includegraphics[width=0.7\columnwidth]{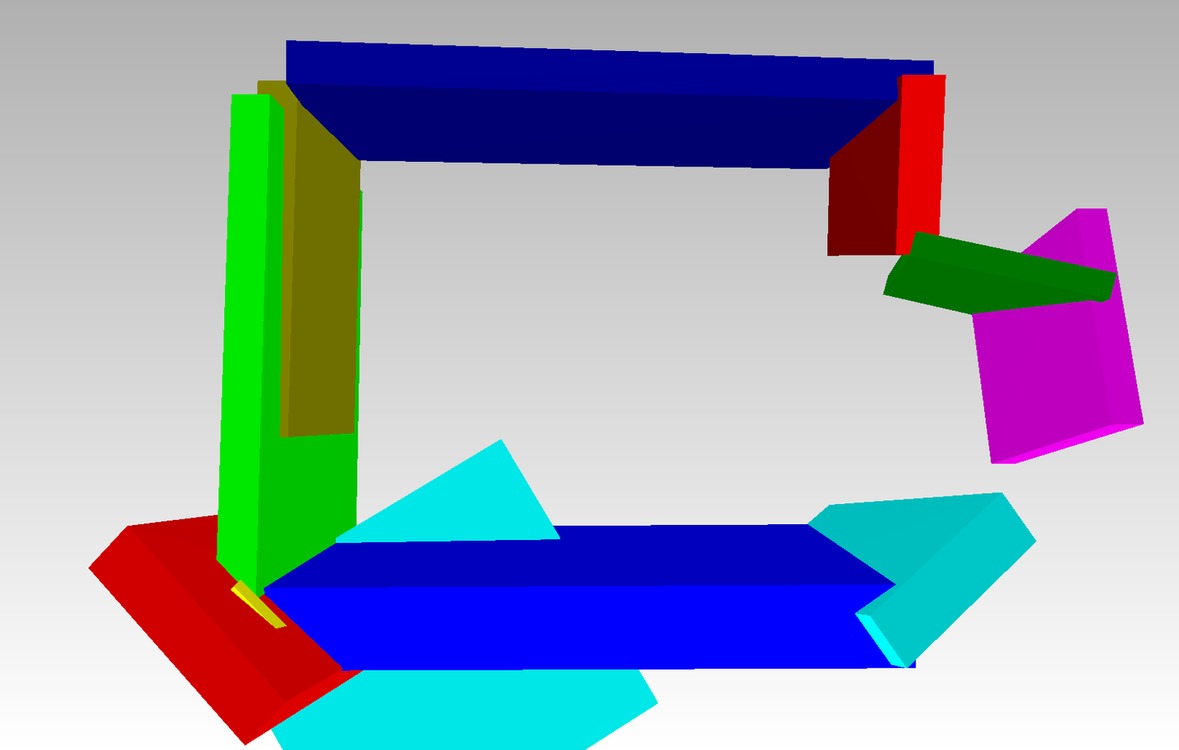}
\end{tabular}
\end{center}
\caption{Reconstructed structure of Castle-P30: points (left), lines (middle), bounding boxes of coplanarity planes after BA (right).}
\label{fig::castle}
\vspace{-2mm}
\end{figure*}

\smallsection{State-of-the-art accuracy}\label{sotaaccur}
Tab.~\ref{tab::overall} shows our general approach outperforms incremental point-based SfM methods \cite{Snavely2006Bundler,Wu2013linsfm} regarding accuracy, and is on par with state-of-the-art global SfM methods \cite{moulon2013global,Olsson2011scia,Cui2015bmvc,ArieNachimson2012global,Jiang2013iccv}.  The wider applicability thus was not traded for accuracy, even in scenarios with dense features and much overlap.

It also shows that our relaxed trifocal constraints retain the most relevant part of trifocality contribution to calibration.  Comparing to Tab.~\ref{tab::ransac_features} (right), we can see that coplanarity constraints alone provide in general comparable or better accuracy than incremental SfM methods \cite{Snavely2006Bundler,Wu2013linsfm}, except again on Fountain-P11.  Residuals are shown in Tab.~\ref{tab::residuals}.


We could not compare with line-based SfM methods \cite{ZHANG_SfM_IJCV14} as their code is unavailable and as they do not measure their accuracy against standard datasets with ground truth.

\smallsection{Succeeding when others fail}
Our method can calibrate difficult datasets that other methods fail to calibrate:

In Office-P19, some triplets do not have overlapping features (neither points nor lines), which makes it impossible to calibrate with existing SfM methods.  Our method succeeds.  Fig.~\ref{fig::coplanar} illustrates some (consecutive) views of the dataset with line detections, and point, line and plane reconstructions.  Apart from a few line outliers due to mismatches, line and plane reconstructions are qualitatively good.  We can also calibrate Meeting-P31 and Trapezoid-P17, whereas OpenMVG fails; the residuals and number of calibrated cameras are given in Tab.~\ref{tab::residuals}.

\begin{figure}[t]
\begin{center}
\includegraphics[width=0.47\columnwidth]{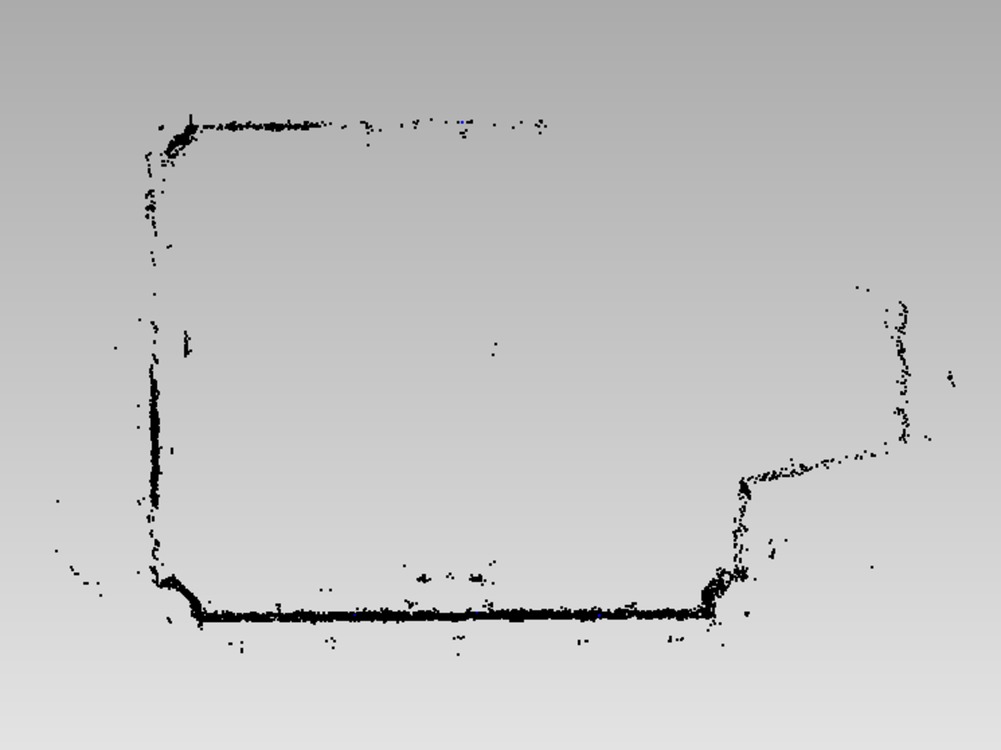} ~\!
\includegraphics[width=0.47\columnwidth]{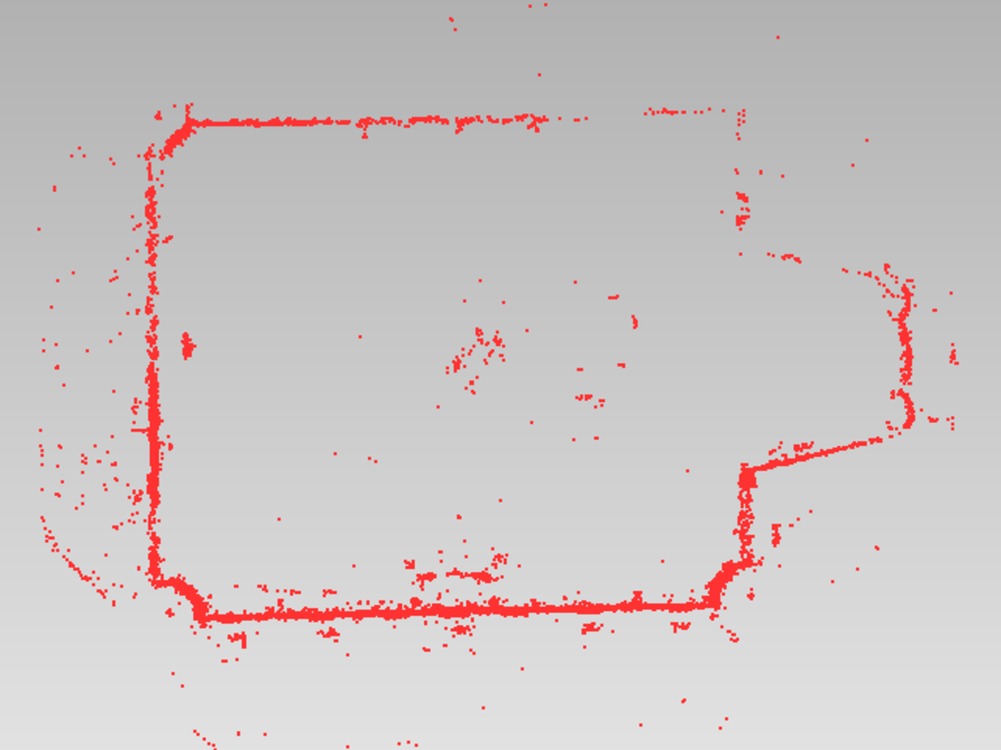}
\end{center}
\vspace{-4mm}
\caption{Castle-P(19$-$3): OpenMVG (left), our method (right).}
\label{fig::castlep16}
\vspace{-1mm}
\begin{center}
	\includegraphics[width=0.3\columnwidth]{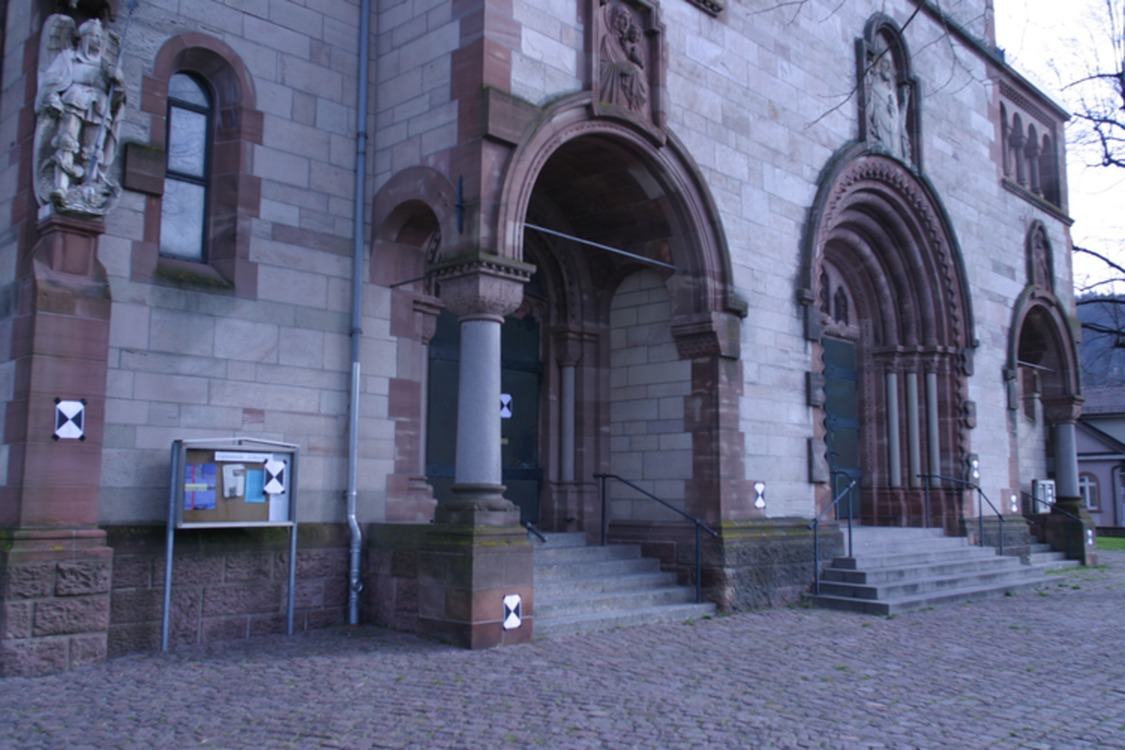} ~
	\includegraphics[width=0.3\columnwidth]{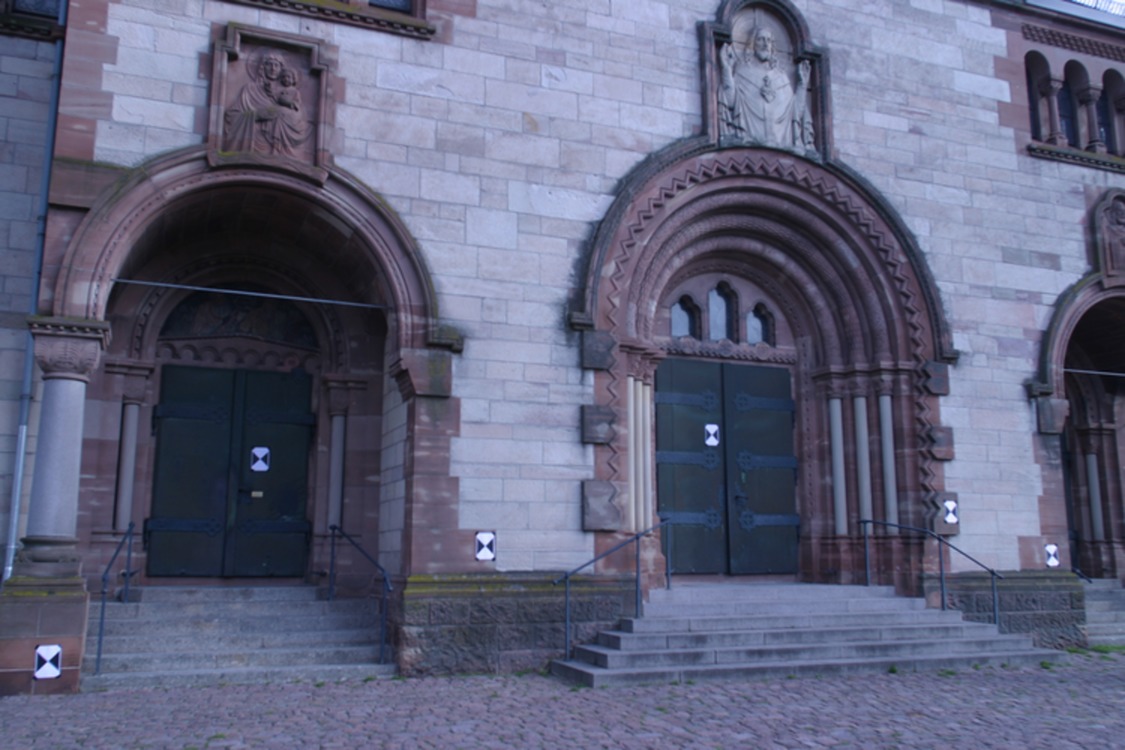} ~
	\includegraphics[width=0.3\columnwidth]{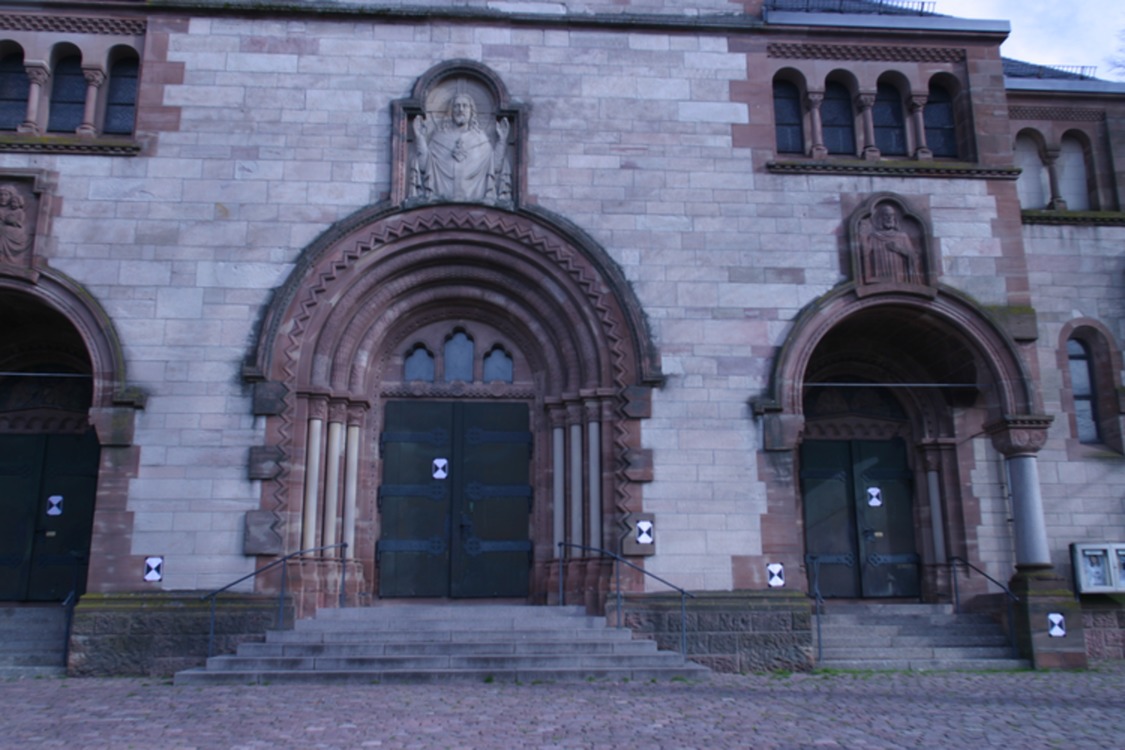} ~
\end{center}
\vspace{-3mm}
\caption{Only our method can calibrate the very wide baseline and viewpoint change of Herz-Jesu-P3 (furthest 3 images in P8).}
\label{fig::herzjesup3}
\vspace{-2mm}
\end{figure}

Removing 3/19 images from Castle-P19 can be enough to cause other methods to fail to calibrate all cameras, thus also reducing reconstructed structures and ability to model the whole scene (cf.\ missing wall parts in Fig.~\ref{fig::castlep16}, left). Our method calibrates all cameras, yielding an average location error of 13.2\,cm. It is not as good as the state-of-the-art error of 2.3\,cm when all 19 images are available (cf.\ Tab.~\ref{tab::overall}), but still better than Bundler (34.4\,cm) and VSfM (25.8\,cm).
We can actually remove up to 11/19 images and still estimate all camera poses with average error 18.4\,cm, which remains better than incremental methods with all 19 images.

Likewise, keeping only 3 images (left-most, middle, right-most) from Herz-Jesu-P8 (Fig.~\ref{fig::herzjesup3}) leads other methods to totally fail, while we calibrate these very-wide-baseline images with 6\,mm accuracy, the same as full P8.

More generally, disconnected components in the trifocal graph make other SfM methods fail to calibrate all cameras (in the same reference frame), even if the bifocal graph is connected.  (For other methods to work, connections in the trifocal graph actually even have to be supported by at least 3 features.)  However, provided there are 3D lines in the configuration described in Fig.~\ref{fig::intro}, we can hope to calibrate all cameras in the dataset.

\section{Conclusion}

We have presented a novel SfM method that exploits lines and coplanarity constraints to estimate camera poses in difficult settings with little image overlap and possibly little or no texture, causing other methods to fail.

Our experiments show that our method combines the accuracy of usual trifocal constraints although we relax them, and the robustness (in the sense of wider applicability) of coplanar constraints. It can thus be blindly used to calibrate scenes that other methods fail to calibrate, and still to provide state-of-the-art results on less difficult scenes, with smaller baselines, more overlap or more texture.


The coplanarity constraint we have defined is actually not intrinsically specific to lines. It could apply to points as well, as long as they can be associated to planes, e.g., using a method to fit multiple homographies in two views \cite{Magri2014cvpr}. 
Future work also include the use of coplanarity constraints for dense surface reconstruction, even in textureless scenes.

{\small
\bibliographystyle{ieee}
\bibliography{egbib}

\begin{thebibliography}{10}\itemsep=-1pt

\bibitem{ceres}
S.~Agarwal, K.~Mierle, and Others.
\newblock Ceres solver.
\newblock \url{http://ceres-solver.org}.

\bibitem{ArieNachimson2012global}
M.~Arie-Nachimson, S.~Z. Kovalsky, I.~Kemelmacher-Shlizerman, A.~Singer, and
  R.~Basri.
\newblock Global motion estimation from point matches.
\newblock In {\em 3DIMPVT}, 2012.

\bibitem{Arrigoni2015transnorm}
F.~Arrigoni, B.~Rossi, and A.~Fusiello.
\newblock On computing the translations norm in the epipolar graph.
\newblock In {\em International Conference on 3D Vision (3DV 2015)}, 2015.

\bibitem{Bartoli2003ijcv}
A.~Bartoli and P.~Sturm.
\newblock Constrained {Structure and Motion} from multiple uncalibrated views
  of a piecewise planar scene.
\newblock {\em International Journal of Computer Vision (IJCV 2003)},
  52(1):45--64, 2003.

\bibitem{Bartoli2005}
A.~Bartoli and P.~F. Sturm.
\newblock Structure-from-motion using lines: Representation, triangulation, and
  bundle adjustment.
\newblock {\em Computer Vision and Image Understanding}, 100(3):416--441, 2005.

\bibitem{Cui2015bmvc}
Z.~Cui, N.~Jiang, C.~Tang, and P.~Tan.
\newblock Linear global translation estimation with feature tracks.
\newblock In {\em British Machine Vision Conference (BMVC 2015)}, 2015.

\bibitem{CVPR11}
A.~Elqursh and A.~Elgammal.
\newblock Line-based relative pose estimation.
\newblock In {\em IEEE Conference on Computer Vision and Pattern Recognition
  (CVPR)}, pages 3049--3056, June 2011.

\bibitem{Enqvist2011iccvw}
O.~Enqvist, F.~Kahl, and C.~Olsson.
\newblock Non-sequential structure from motion.
\newblock In {\em ICCV Workshops}, pages 264--271, 2011.

\bibitem{Frahm2010rome}
J.-M. Frahm, P.~Fite-Georgel, D.~Gallup, T.~Johnson, R.~Raguram, C.~Wu, Y.-H.
  Jen, E.~Dunn, B.~Clipp, S.~Lazebnik, and M.~Pollefeys.
\newblock Building {Rome} on a cloudless day.
\newblock In K.~Daniilidis, P.~Maragos, and N.~Paragios, editors, {\em 11th
  European Conference on Computer Vision (ECCV 2010)}, pages 368--381. Springer
  Berlin Heidelberg, 2010.

\bibitem{Fusiello2015npnp}
A.~Fusiello, F.~Crosilla, and F.~Malapelle.
\newblock Procrustean point-line registration and the {NPnP} problem.
\newblock In {\em International Conference on 3D Vision (3DV 2015)}, pages
  250--255, Oct. 2015.

\bibitem{Garro20123DV}
V.~Garro, F.~Crosilla, and A.~Fusiello.
\newblock Solving the {PnP} problem with anisotropic orthogonal {Procrustes}
  analysis.
\newblock In {\em International Conference on 3D Imaging, Modeling, Processing,
  Visualization Transmission (3DIMPVT 2012)}, pages 262--269, Oct. 2012.

\bibitem{Govindu2001cvpr}
V.~M. Govindu.
\newblock Combining two-view constraints for motion estimation.
\newblock In {\em Conference on Computer Vision and Pattern Recognition (CVPR
  2001)}, 2001.

\bibitem{Govindu2006robustness}
V.~M. Govindu.
\newblock Robustness in motion averaging.
\newblock In {\em ACCV}, 2006.

\bibitem{Havlena2010eccv}
M.~Havlena, A.~Torii, and T.~Pajdla.
\newblock Efficient {Structure from Motion} by graph optimization.
\newblock In {\em 11th European Conference on Computer Vision (ECCV 2010)},
  pages 100--113, Berlin, Heidelberg, 2010.

\bibitem{Heinly2015cvpr}
J.~Heinly, J.~L. Sch\"onberger, E.~Dunn, and J.-M. Frahm.
\newblock Reconstructing the world* in six days *(as captured by the {Yahoo}
  100 million image dataset).
\newblock In {\em Conference on Computer Vision and Pattern Recognition (CVPR
  2015)}, 2015.

\bibitem{Jiang2013iccv}
N.~Jiang, Z.~Cui, and P.~Tan.
\newblock A global linear method for camera pose registration.
\newblock In {\em {IEEE} International Conference on Computer Vision (ICCV
  2013)}, pages 481--488, 2013.

\bibitem{Jiang2015cvpr}
N.~Jiang, W.~Lin, M.~N. Do, and J.~Lu.
\newblock Direct structure estimation for {3D} reconstruction.
\newblock In {\em {IEEE} Conference on Computer Vision and Pattern Recognition
  (CVPR 2015)}, pages 2655--2663, 2015.

\bibitem{KahlH2008pami}
F.~Kahl and R.~I. Hartley.
\newblock Multiple-view geometry under the $l_\infty$-norm.
\newblock {\em IEEE Trans. PAMI}, 30(9):1603--1617, 2008.

\bibitem{Kim2014accv}
C.~Kim and R.~Manduchi.
\newblock Planar structures from line correspondences in a {Manhattan} world.
\newblock In {\em 12th Asian Conference on Computer Vision (ACCV 2014)}, pages
  509--524, 2014.

\bibitem{Lepetit2008ijcv}
V.~Lepetit, F.~Moreno-Noguer, and P.~Fua.
\newblock {EPnP}: An accurate {O(n)} solution to the {PnP} problem.
\newblock {\em International Journal of Computer Vision (IJCV 2008)},
  81(2):155--166, 2008.

\bibitem{SIFT}
D.~G. Lowe.
\newblock Distinctive image features from scale-invariant keypoints.
\newblock {\em Int. J. Comput. Vision}, 60(2):91--110, Nov. 2004.

\bibitem{Magri2014cvpr}
L.~Magri and A.~Fusiello.
\newblock T-linkage: {A} continuous relaxation of {J}-linkage for multi-model
  fitting.
\newblock In {\em {IEEE} Conference on Computer Vision and Pattern Recognition
  (CVPR 2014)}, pages 3954--3961, 2014.

\bibitem{Mirzaei2011pnl}
F.~M. Mirzaei and S.~I. Roumeliotis.
\newblock Globally optimal pose estimation from line correspondences.
\newblock In {\em IEEE International Conference on Robotics and Automation
  (ICRA 2011)}, pages 5581--5588. IEEE, 2011.

\bibitem{ACRANSAC}
L.~Moisan and B.~Stival.
\newblock A probabilistic criterion to detect rigid point matches between two
  images and estimate the fundamental matrix.
\newblock {\em Int. J. Comput. Vision}, 57(3):201--218, May 2004.

\bibitem{ACPOINTS}
P.~Moulon, P.~Monasse, and R.~Marlet.
\newblock Adaptive {Structure from Motion} with a contrario model estimation.
\newblock In {\em 11th Asian Conference on Computer Vision (ACCV 2012) - Volume
  Part IV}, ACCV'12, pages 257--270, Berlin, Heidelberg, 2013. Springer-Verlag.

\bibitem{moulon2013global}
P.~Moulon, P.~Monasse, and R.~Marlet.
\newblock Global fusion of relative motions for robust, accurate and scalable
  structure from motion.
\newblock In {\em Proceedings of the IEEE International Conference on Computer
  Vision (ICCV)}, pages 3248--3255, 2013.

\bibitem{Olsson2011scia}
C.~Olsson and O.~Enqvist.
\newblock Stable structure from motion for unordered image collections.
\newblock In {\em SCIA}, pages 524--535, 2011.
\newblock LNCS 6688.

\bibitem{Ozyesil2015cvpr}
O.~\"Ozye\c{s}il and A.~Singer.
\newblock Robust camera location estimation by convex programming.
\newblock In {\em IEEE Conference on Computer Vision and Pattern Recognition
  (CVPR 2015)}, June 2015.

\bibitem{Pribyl2015bmvc}
B.~P\v{r}ibyl, P.~Zem\v{c}\'{\i}k, and M.~\v{C}adik.
\newblock Camera pose estimation from lines using {P}lücker coordinates.
\newblock In X.~Xie, M.~W. Jones, and G.~K.~L. Tam, editors, {\em British
  Machine Vision Conference (BMVC 2015)}, pages 45.1--45.12. BMVA Press, Sept.
  2015.

\bibitem{Ramalingam2011icra}
S.~Ramalingam, S.~Bouaziz, and P.~F. Sturm.
\newblock Pose estimation using both points and lines for geo-localization.
\newblock In {\em {IEEE} International Conference on Robotics and Automation
  (ICRA 2011)}, pages 4716--4723, 2011.

\bibitem{Rodriguez2011cvpr}
A.~L. Rodr\'{\i}guez, P.~E. L{\'o}pez-de Teruel, and A.~Ruiz.
\newblock Reduced epipolar cost for accelerated incremental {SfM}.
\newblock In {\em Conference on Computer Vision and Pattern Recognition (CVPR
  2011)}, 2011.

\bibitem{Rother2003iccv}
C.~Rother.
\newblock Linear multi-view reconstruction of points, lines, planes and cameras
  using a reference plane.
\newblock In {\em 9th IEEE International Conference on Computer Vision (ICCV
  2003) - Volume 2}, ICCV '03, pages 1210--1217, Washington, DC, USA, 2003.
  IEEE Computer Society.

\bibitem{multiLSD}
Y.~Sala\"un, R.~Marlet, and P.~Monasse.
\newblock Multiscale line segment detector for robust and accurate {SfM}.
\newblock In {\em 23rd International Conference on Pattern Recognition (ICPR)},
  2016.

\bibitem{ECCV16}
Y.~Sala\"un, R.~Marlet, and P.~Monasse.
\newblock Robust and accurate line- and/or point-based pose estimation without
  {M}anhattan assumptions.
\newblock In {\em European Conference on Computer Vision (ECCV)}, 2016.

\bibitem{Sharp01towardmultiview}
G.~C. Sharp, S.~W. Lee, and D.~K. Wehe.
\newblock Toward multiview registration in frame space.
\newblock In {\em IEEE International Conference on Robotics and Automation
  (ICRA 2001)}, 2001.

\bibitem{Sim2006cvpr}
K.~Sim and R.~Hartley.
\newblock Recovering camera motion using $l_\infty$ minimization.
\newblock In {\em Conference on Computer Vision and Pattern Recognition (CVPR
  2006)}, volume~1, 2006.

\bibitem{Snavely2006Bundler}
N.~Snavely, S.~M. Seitz, and R.~Szeliski.
\newblock Photo tourism: Exploring photo collections in {3D}.
\newblock {\em ACM Transations on Graphics (TOG 2006)}, 25(3):835--846, July
  2006.

\bibitem{STRECHA}
C.~Strecha, W.~von Hansen, L.~Van~Gool, P.~Fua, and U.~Thoennessen.
\newblock On benchmarking camera calibration and multi-view stereo for high
  resolution imagery.
\newblock In {\em IEEE Conference on Computer Vision and Pattern Recognition
  (CVPR 2008)}, pages 1--8, June 2008.

\bibitem{Sturm2000cvpr}
P.~F. Sturm.
\newblock Algorithms for plane-based pose estimation.
\newblock In {\em Conference on Computer Vision and Pattern Recognition (CVPR
  2000 )}, pages 1706--1711, 2000.

\bibitem{Sweeney2015iccv}
C.~Sweeney, T.~Sattler, T.~Hollerer, M.~Turk, and M.~Pollefeys.
\newblock Optimizing the viewing graph for {Structure-from-Motion}.
\newblock In {\em IEEE International Conference on Computer Vision (ICCV
  2015)}, pages 801--809, Washington, DC, USA, 2015.

\bibitem{Toldo2015cviu}
R.~Toldo, R.~Gherardi, M.~Farenzena, and A.~Fusiello.
\newblock Hierarchical structure-and-motion recovery from uncalibrated images.
\newblock {\em Computer Vision and Image Understanding (CVIU 2015)},
  140(C):127--143, Nov. 2015.

\bibitem{Wilson2014eccv}
K.~Wilson and N.~Snavely.
\newblock Robust global translations with {1DSfM}.
\newblock In D.~Fleet, T.~Pajdla, B.~Schiele, and T.~Tuytelaars, editors, {\em
  13th European Conference on Computer Vision (ECCV 2014)}, pages 61--75, 2014.

\bibitem{Wu2013linsfm}
C.~Wu.
\newblock Towards linear-time incremental structure from motion.
\newblock In {\em International Conference on 3D Vision (3DV 2013)}, pages
  127--134, 2013.

\bibitem{Xu2016pnl}
C.~Xu, L.~Zhang, L.~Cheng, and R.~Koch.
\newblock Pose estimation from line correspondences: A complete analysis and a
  series of solutions.
\newblock {\em IEEE Transactions on Pattern Analysis and Machine Intelligence
  (PAMI 2016)}, June 2016.

\bibitem{ZachK2010cvpr}
C.~Zach, M.~Klopschitz, and M.~Pollefeys.
\newblock Disambiguating visual relations using loop constraints.
\newblock In {\em Conference on Computer Vision and Pattern Recognition (CVPR
  2010)}, 2010.

\bibitem{LBD}
L.~Zhang and R.~Koch.
\newblock An efficient and robust line segment matching approach based on {LBD}
  descriptor and pairwise geometric consistency.
\newblock {\em J. Vis. Comun. Image Represent.}, 24(7):794--805, Oct. 2013.

\bibitem{ZHANG_SfM_IJCV14}
L.~Zhang and R.~Koch.
\newblock Structure and motion from line correspondences: Representation,
  projection, initialization and sparse bundle adjustment.
\newblock {\em Journal of Visual Communication and Image Representation},
  25(5):904--915, 2014.

\bibitem{Zhang2012accv}
L.~Zhang, C.~Xu, K.-M. Lee, and R.~Koch.
\newblock Robust and efficient pose estimation from line correspondences.
\newblock In K.~M. Lee, Y.~Matsushita, J.~M. Rehg, and Z.~Hu, editors, {\em
  11th Asian Conference on Computer Vision (ACCV 2012)}, pages 217--230.
  Springer, 2012.

\bibitem{Zheng2013iccv}
Y.~Zheng, Y.~Kuang, S.~Sugimoto, K.~\AA{}str\"{o}m, and M.~Okutomi.
\newblock Revisiting the {PnP} problem: A fast, general and optimal solution.
\newblock In {\em IEEE International Conference on Computer Vision (ICCV
  2013)}, pages 2344--2351, Dec. 2013.

\bibitem{Zhou2012cvpr}
Z.~Zhou, H.~Jin, and Y.~Ma.
\newblock Robust plane-based {Structure from Motion}.
\newblock In {\em IEEE Conference on Computer Vision and Pattern Recognition
  (CVPR 2012)}, pages 1482--1489, Washington, DC, USA, 2012.

\end{thebibliography}
}

\end{document}